\documentclass[screen]{acmart}

\usepackage{svg}
\usepackage{balance}
\usepackage{subcaption}
\usepackage{graphicx}
\usepackage{multirow}
\usepackage{hyperref}
\usepackage{colortbl}

\usepackage[]{algpseudocode}
\usepackage{algorithm}
\algnewcommand\algorithmicforeach{\textbf{for each}}
\algdef{S}[FOR]{ForEach}[1]{\algorithmicforeach\ #1\ \algorithmicdo}
\usepackage{arydshln}
\usepackage{flushend}

\newcommand{\Real}{\mathop{\rm I\kern-.2emR}}

\AtBeginDocument{%
  }

\setcopyright{acmcopyright}

\acmJournal{JDS}
\acmVolume{37}
\acmNumber{4}
\acmArticle{111}
\acmMonth{8}

\begin{document}

\title{Multiobjective Evolutionary Component Effect on Algorithm behavior}

\author{Yuri Lavinas}
\email{lavinas.yuri.xp@alumni.tsukuba.ac.jp}
\orcid{0000-0003-2712-5340}
\affiliation{%
  \institution{IRIT - CNRS UMR5505, University of Toulouse}
  \country{France}
}

\author{Marcelo Ladeira}
\orcid{0000-0003-1542-6293}
\email{mladeira@unb.br}
\affiliation{%
  \institution{University of Brasilia}
  \country{Brazil}
}

\author{Gabriela Ochoa}
\orcid{0000-0001-7649-5669}
\email{gabriela.ochoa@cs.stir.ac.uk}
\affiliation{%
  \institution{University of Stirling}
  \country{UK}
}

\author{Claus Aranha}
\orcid{0000-0003-1390-7536}
\email{caranha@cs.tsukuba.ac.jp}
\affiliation{%
  \institution{University of Tsukuba}
  \country{Japan}
}

\acmArticleType{Research}
\acmCodeLink{https://github.com/yurilavinas/MOEADr/tree/telo22}
\acmDataLink{https://doi.org/10.5281/zenodo.8192256}

\begin{abstract}
   The performance of multiobjective evolutionary algorithms (MOEAs) varies across problems, making it hard to develop new algorithms or apply existing ones to new problems. To simplify the development and application of new multiobjective algorithms, there has been an increasing interest in their automatic design from their components. These automatically designed metaheuristics can outperform their human-developed counterparts. However, it is still unknown what are the most influential components that lead to performance improvements. This study specifies a new methodology to investigate the effects of the final configuration of an automatically designed algorithm. We apply this methodology to a tuned Multiobjective Evolutionary Algorithm based on Decomposition (MOEA/D) designed by the iterated racing (irace) configuration package on constrained problems of 3 groups: (1) analytical real-world problems, (2) analytical artificial problems and (3) simulated real-world. We then compare the impact of the algorithm components in terms of their Search Trajectory Networks (STNs), the diversity of the population, and the anytime hypervolume values. Looking at the objective space behavior, the MOEAs studied converged before half of the search to generally good HV values in the analytical artificial problems and the analytical real-world problems. For the simulated problems, the HV values are still improving at the end of the run. In terms of decision space behavior, we see a diverse set of the trajectories of the STNs in the analytical artificial problems. These trajectories are more similar and frequently reach optimal solutions in the other problems. 

\end{abstract}

\ccsdesc[500]{Computing methodologies~Continuous space search}

\keywords{algorithm analysis, continuous optimization, automatic algorithm configuration, multiobjective optimization}

\maketitle

\section{Introduction}
\label{section:intro}

Multiobjective Optimization Problems (MOPs) are problems with two or more conflicting objective functions that are optimized simultaneously. Several multiobjective evolutionary algorithms (MOEAs) have been proposed to solve MOPs~\cite{deb2002fast,beume2007sms,ZhaLi07:moead}, and for each of these algorithms several variants also have been designed, where some of the components of the algorithms are modified.

Recently, there has been an increasing interest in the automatic design of MOEAs~\cite{Radulescu2013,jmetal_irace,moead_irace}. In these approaches, a configurator recombines components of established MOEAs, creating more effective variants. However, the specific reasons for these improvements are  still not clear. 
We argue that by understanding how specific components contribute to the performance of the variants, we can develop new and better components.

In this work, we investigate the question of how to measure the contribution of specific components in a MOEA. We focus this investigation on understanding the changes of \emph{anytime behavior} effected on the algorithm by one or more components. We define anytime behavior as the grouping of the algorithm's performance in terms of objective space convergence, and the choices made during the exploration of the decision space. We highlight that this definition focuses on the behavior of the algorithm when optimizing a problem, thus dissociating from \emph{landscape analysis}, which focuses on the structure of the search space.




To analyze the contributions of specific components, we perform a case study on a variant of the MOEA/D algorithm~\cite{ZhaLi07:moead} created by an algorithm configurator (irace). The MOEA/D is a popular and efficient algorithm for solving MOPs and can modify its search behavior on different MOPs. 

To understand the contribution of each component, we perform modifications of the above automatically designed MOEA/D (auto-MOEA/D), by either removing a component, or replacing the component with the corresponding original in MOEA/D, as appropriate. For each modification, we investigate its performance on a set of problems to identify the contribution of the component. The methodology is an extension of our work introduced in~\cite{lavinas_gecco2022}, with the inclusion of objective space behavior analysis in the form of anytime hypervolume analysis.




This investigation takes the form of a case study on six real-world analytical continuous benchmark problems, compiled together by Tanabe et. al~\cite{TANABE2020} and two simulated continuous benchmark problems: (1) the problem of selecting landing sites for a lunar exploration robot~\cite{MoonOrbitingSatellite2015} and (2) the problem of optimization of car designs~\cite{kohira2018proposal}. We conduct our analysis focusing on how these metaheuristics explore both the objective and decision space. Furthermore, we contrast the automatically designed MOEA (auto-MOEA/D) against each of the variants in terms of their Search Trajectory Networks (STNs)~\cite{stn_main,STN_MOP_evostar}; the diversity of their populations and the traditional performance metrics. Moreover, we compare the analytical and simulated problems in terms of the overall metrics of auto-MOEA/D. Finally, we analyze the similarities between the benchmark problems used for designing auto-MOEA/D and the real-world problems. To the best of our knowledge, this is the first component-wise analysis of MOEAs on the objective and decision space dynamics in real-world constrained MOPs. For reproducibility purposes, all the code and experimental scripts are available online at \href{https://doi.org/10.5281/zenodo.8192256}{https://doi.org/10.5281/zenodo.8192256}. In this paper, our contributions can be summarised as follows:%

\begin{itemize}
    \item[(1)] We extend our previous study on behavior analysis to a more thorough investigation, by considering the differences in behavior in both the objective space and decision space during the search progress.
    \item[(2)] We study the behavior of the components of MOEA/D in analytical and simulated real-world problems.
    \item[(3)] We analyze the similarities between the benchmark problems used for designing auto-MOEA/D and the real-world problems. 
\end{itemize}

The paper is organized as follows. Section~\ref{related-work} overviews previous work related to the automated design of algorithms and constrained problems. Section~\ref{prelim} introduces relevant concepts followed by Section~\ref{beheaviour_method} that explains the details of the methodology used in this work. The automatic design of the MOEA is shown in Section~\ref{exp_automatic}. Then, the comparison of the components set-up is presented in Section~\ref{search_behavior}, and the analysis of the search behaviors dynamics of the different MOEA/D variants is shown in Section~\ref{dynamics}. Finally, Section~\ref{discussion} outlines our main findings, limitations and suggestions for future work.

\section{Related Work}\label{related-work}

\subsection{Analysis of Algorithm Behavior}

Convergence analysis~\cite{98Eiben} is a way to describe algorithm behavior, by illustrating the trade-off between exploration and exploitation in evolutionary algorithms.
However, the information of whether a population has converged or not does not inform the location of this convergence, hence it does not allow the user to know whether the convergence is premature or not. In this way, understanding the behavior of search and optimization algorithms remains a challenge.
%

Another way to understand the behavior of an algorithm, especially in the case of multiobjective optimization, is to visualize and contrast the Pareto front achieved by the algorithm~\cite{fonseca1996performance,performance_assessment,KerschkeG17,OnePLOTtoShow,Liefooghe2018OnPL,Fieldsend2019VisualisingTL}. However, in general this approach focuses on detecting increments in performance, and limits to observing changes in the \emph{objective space dynamics}. 
We argue that analysing the decision space might expand our understanding of the behavior of the multiobjective optimisation solvers. 

Finally, another way to understand algorithm behavior is through Search Trajectory Networks (STNs)~\cite{stn_evostar,stn_main}, which illustrate as a graph how the algorithm explores the decision space. Recently, STNs were generalized to multiobjective algorithms~\cite{STN_MOP_evostar}, and we use these MOP-STN models as one of the tools to discriminate behavior changes in MOEA components.

\subsection{Automatic Design of Evolutionary Algorithms}

Most approaches to the automatic design of evolutionary algorithms focus on creating templates that can instantiate many algorithms and their parameter settings for performance improvements. For example, there have been studies to automatically design NSGA-II~\cite{jmetal_irace} and MOEA/D~\cite{moead_irace} on commonly used benchmark sets. Moreover, two seminal examples are the works of Bezerra et al.~\cite{bezerra2016automatic,Bezerra2020}, which proposed a component-wise MOEA template that instantiates multiple existing frameworks for continuous and combinatorial optimisation MOPs.  Their research efforts mainly focus on exploiting the automatic configuration to increase the performance of multiobjective algorithms in benchmark problems without constraints. 

We also highlight the work of Radulescu et al.~\cite{Radulescu2013}, which focuses on improving the performance of multiobjective metaheuristics. These works are insightful approaches; however, they concentrate on finding well-performing configurations of multiobjective algorithms. On the other hand, there are few studies in the context of the automatic design of algorithms that focus on \emph{the effect of the different components} on the performance of the algorithm.

\subsection{Behavior Analysis in the Continuous Multiobjective Domain}

There are few works in the continuous multiobjective domain related to behavior analysis. Some of the few examples with works on the relations between behavior and population size \cite{behavior_popsize_continuous} and between behavior and solution quality and time \cite{radulescu2013automatically}. In contrast more works studied the behavior of multiobjective algorithms in the combinatorial domain. Two of these works focus on understanding selection and population size effects on the algorithms’ ability in terms of dominance status, membership to the Pareto optimal set, recentness of discovery, and how their numbers change generation by generation~\cite{app_resolution_aguirre,metrics_resolution_aguirre}. Another work explores how the relationships of already known and controllable structures, such as modality and ruggedness to understand the working principles, behavior, and the performance of MOEAs~\cite{epistatic_aguirre}. Finally, few studies have considered the contribution of individual components to MOEAs performance~\cite{bezerra2016automatic}. Furthermore, in most cases, performance is evaluated in unconstrained problems or problems where constraints are simple to address~\cite{TANABE2020,bezerra_ecj,felipe_CHT}. These constraints invalidate some solutions, which makes finding a set of feasible solutions a challenging task.

\section{Preliminaries}\label{prelim}

MOEA/D~\cite{ZhaLi07:moead} is a popular and efficient algorithm for finding good sets of trade-off solutions for MOPs. The key idea of MOEA/D is to create a linear decomposition of the MOP into a set of single-objective subproblems. Decomposing the MOP in various single-objective subproblems makes the algorithm very flexible for dealing with constraints because adding a penalty value is straightforward: MOEA/D adds a penalty value related to the amount of violation of the constraint for each of the subproblems. Given the nature of the single-objective subproblems, MOEA/D can easily use multiple constraint handling techniques (CHTs).


\begin{algorithm}[t]
	\caption{MOEA/D outline}\label{algo:moead}
	\begin{algorithmic}[1]
		\State Initialize population and decomposition vectors.
		
		\While{\textit{Computational budget is not meet}}
		    
            \State \textbf{Define} Neighbourhood relations.
		    
		    \State \textbf{If} partial update is used, select subset of solutions to update. 
		    \State \textbf{Else}, all solutions are updated.
            
            \State \textbf{Generate} candidates from the updated solutions and their neighbors.
            \State \textbf{Evaluate} the candidates on their respective subproblems.
            
            \State \textbf{Update} the population using the candidates.
            \State \textbf{Save} non-dominated solutions in UEA.
            
		    \State \textbf{if} restart criteria is met, regenerate the population.
		\EndWhile
	\end{algorithmic}
\end{algorithm}

The MOEA/D template we propose for instantiating and
designing variants of this metaheuristic is shown in Algorithm~\ref{algo:moead}. We use the generational version of MOEA/D incremented with the Unbounded External Archive (UEA). The UEA is used to keep all nondominated solutions found by a multiobjective optimizer during the search process. Solutions in the archive are only used as the output of the algorithm and are stored in a way that they do not affect the search run~\cite{uea2, tanabe2017benchmarking}.

\subsection{Automatic Design Configurator}

For automated design, we use Iterated Racing (irace)~\cite{LOPEZIBANEZ201643}. The goal of using irace is to be able to tune the set of components of an algorithm over a set of optimization problems to find a configuration that performs well on average in all problems. After fine-tuning the MOEA/D with irace, we conduct an ablation analysis~\cite{fawcett2016analysing, LOPEZIBANEZ201643} to help us understand the choice of components values and whether each of these choices effectively improves the MOEA/D performance. This analysis investigates the differences between configurations. We conduct an ablation analysis between a target configuration selected~\footnote{In our case, we select the first configuration tried by irace during the tuning process.} and the best configuration found by irace.

\subsection{Search Trajectory Networks (STNs) for MOPs}\label{STNS_mops}

We use Search Trajectory Networks as a tool for visualization following the method described in~\cite{STN_MOP_evostar, lavinas_gecco2022}. In an STN model, each solution in the decision space is mapped to a location. Similar solutions are generally mapped to the same location, as the locations represent a partition of the decision space. The network models are extracted from data obtained during several runs of the studied algorithm(s). A network model requires defining its nodes and edges. In an STN model, \textit{nodes} are the locations in the search trajectories visited by a given algorithm, and \textit{edges} connects two consecutive locations in the trajectory. A strength of network models is that they can be visualized. When decorating the networks for visualization, it is possible to highlight attributes of the nodes and edges that are relevant to the search process. In these visualizations, the size of nodes and the width of edges are proportional to how many times the algorithms visited them during the aggregation of runs used to extract the model. Visualizations use \emph{force-directed} graph layout algorithms as implemented in \textsf{R} package igraph \cite{igraph}.

The key idea of this method is that we keep track of a small number of decomposition vectors, match a representative solution to a vector, and then merge the vector trajectories of each vector into a single multiobjective STN. The merged STN model merges the $n$ STNs of each decomposition vector and is obtained by the graph union of the $n$ individual graphs. The merged graph contains the nodes and edges present in at least one of the vectors graphs. Attributes are kept for the nodes and edges, indicating whether they were visited by both algorithms (shared) or by one of them only. Finally, we have the merged STN models, where different MOEAs are combined into one single merged STN model. The merged STNs allow us to directly visually compare how distinct variants explore the decision space~\cite{lavinas_gecco2022}. In the merged models, there is the notion of \textit{shared nodes}, which are nodes visited by more than one algorithm and are indicated in grey colour in the network visualization.

\subsection{Network and Performance Metrics}

We use the following STN metrics to assess the global structure of the trajectories and bring insight into the behavior of the MOEAs modelled. These metrics are (1) the number of unique nodes, (2) the number of unique edges and (3) the number of shared nodes between vectors and (4) the number of solutions in the Pareto front. 

For reference, we use the following criteria to compare the results of the different strategies, based on the metric analysis done in the work of~\cite{lavinas_gecco2022}: final approximation hypervolume (HV), the volume of the n-dimensional polygon formed by reference points and solutions. It is worth noting that additional network and MOP metrics could also be considered. These metrics are summarised in Table~\ref{stn_metrics}. We also use the population variance metric.

\begin{table}[htbp]
\centering
\caption{Description of decision space metrics}
\label{stn_metrics}
\begin{tabular}{c|c}
\rowcolor[gray]{.85}Metric & Description \\ 
    Nodes & Total number of nodes, which corresponds to the number of unique locations visited.\\
    \multirow{1}{*}{Edges} &  Total number of edges, the number of unique search transitions.\\
    \multirow{1}{*}{Variance} &          Dispersion of the population in the decision space.\\
    \multirow{1}{*}{\#PF} &          Number of solutions in the theoretical Pareto front or in the best approximation to the Pareto front.
\end{tabular}
\vspace{-0.5em}
\end{table}

\section{behavior Analysis Methodology}\label{beheaviour_method}

Here we define a methodology to measure the differences in decision and objective space dynamics of variants of an algorithm that alter a single component from the base algorithm. Our reason for using such methodology is to identify the most influential components of an algorithm and how they affect the search space dynamics in different problems. Thus, we can identify better the influence of the considered component, even if such method explores a reduced part of the possible algorithm versions.

First, we use irace to automatically design a tailor-suited MOEA (auto-MOEA) that performs well in a set of problems. Secondly, we modify auto-MOEA to create many variants and each one of these variants have only one component that differs from auto-MOEA. These variants are obtained by: (1) removing a component if possible; (2) otherwise, replacing this component with the corresponding one from the original MOEA. Thus, we have the multiple MOEAs: the auto-MOEA plus one variant for each component. The idea here is to be able to capture the effects of each one of the components individually. This step to create variants is done manually, based on the users expertise about the components.

Then, we collect log data of the executions of all variants and the base algorithm during their run in the problems of interest. This log data contains the parameters of the non-dominated solutions of each generation as well as their objective space values and the solution's feasibility~\footnote{For technical reasons we also keep the execution number.}. This data is processed following the approach described in Subsection~\ref{STNS_mops} for analysing how the components affect the decision space exploration of such MOEA. To measure the different behaviors, we use metrics and visuals from the Search Trajectory Networks (STNs)~\cite{stn_main,STN_MOP_evostar} combined with a population diversity metric.

We also use the log data to calculate the anytime HV performance of the algorithm over the evaluations and the number of solutions in the Pareto front\footnote{We use the theoretical Pareto front if it is available. Otherwise, we use the approximation to the Pareto front.}. To verify the impact of the variant, we calculate the difference value of the metrics of each variant to the auto-MOEA.


\section{Designing auto-MOEA/D}
\label{exp_automatic}

We analyse the components of a MOEA/D instance that was automatically designed. This design process was done in a component-wise framework, similar to the protocols used by Bezerra et al.~\cite{bezerra2016automatic} and Campelo et al. ~\cite{moeadr_paper}. We extend the MOEADr package~\cite{moeadr_package} to introduce options for population restart and the most representative Resource Allocation (RA) method, called the partial update of the population ~\cite{lavinas2020moea,lavinas2022faster}. 

\subsection{Variable Components Search Space}

The configuration space used in our experiments contains the algorithm components and numerical parameters of the MOEA/D framework. These are shown in Table~\ref{automated_parameter_table}. Special attention is required with the variation operators: Differential Evolution (DE) mutation and polynomial mutation. They are always performed sequentially, first DE and then the polynomial mutation. Thus, the order of the stack of operators is kept fixed, but the parameter values are variable. Similar attention should be given to the restart strategy, where only the choice of using this strategy is explored. 

\begin{table}[t]

\centering
	\small
	\caption{Components search space.}
	\label{automated_parameter_table}
    \begin{tabular}{l|l}
        
        \rowcolor[gray]{.85}Component            & Domain            \\ 
        \multirow{1}{*}{Decomposition vector generator}    & Uniform or  Sobol\cite{zapotecas-martinezLowdiscrepancySequencesTheir2015} \\ \hline
        \multirow{1}{*}{Population size}         & 100 or  500     \\ \hline    
        \multirow{2}{*}{Aggregation function}         & Weighted Tchebychef or\\
                                                      & Adjusted weighted Tchebychef\\ \hline    
        \multirow{2}{*}{Update strategy}                & Best, $nr = [1, 20]$   \\
                                                        & Restricted, $Tr = [4, 20]$  \\ \hline    
        \multirow{2}{*}{Neighbourhood function} & $T  = [10, 100]$\\    
                                                      & $Delta = [0.1, 1]$ \\ \hline   
        DE mutation                                  & $F = [0.1, 1]$       \\ \hline
        \multirow{2}{*}{Polynomial mutation}        & $\eta_m = [1, 100]$    \\
                                                            & $prob = [0, 1]$     \\ \hline
        \multirow{2}{*}{RA (Partial update strategy)}        & True $n = {0.10, 0.15, 0.20, 0.25}$ \\
                                                        & False, not used\\ \hline  
        \multirow{2}{*}{Restart strategy}        & True, every $20000$ evaluations  \\
                                        & False, not used 
        \end{tabular}
\end{table}


We choose to fix some MOEA/D components to reduce the search space for the irace configurator. The fixed components are the computational budget, the objective scaling and the constraint handling technique (CHT). These fixed components are always present in every configuration of the MOEA/D that irace generates: (1) the number of functions evaluations is set to $100000$ in order to grasp all possible behaviors of the automatically designed algorithm during the run; (2) all objectives were linearly scaled at every iteration to the interval $\left[0,1\right]$; (3) we use the Dynamic CHT~\cite{joines1994use}, which starts with a small penalty value, increases it across the iterations to focus on the diversity of feasible solutions, and then later focus on the convergence of those solutions. It is defined by $f^{agg}_{penalty}(x)  = f^{agg} + (C*t)^\alpha*v(x)$, where $C = 5$ and $\alpha = 2$ are constants we defined based on the following works~\cite{joines1994use,felipe_CHT}, $t$ is the generation number and $v$ is the total violation of a solution.


\subsection{Configurator Setup}

We use the irace configurator~\cite{LOPEZIBANEZ201643} to automatically assemble and design a MOEA/D configuration based on the components available in the MOEADr package extended with the options for population handling mentioned above. We run irace with its default settings, except for the number of elite configurations tested, which we increase from 1 to 7, following Campelo et al. work~\cite{moeadr_paper}. We run irace with a budget of $15 000$ runs.

\subsection{Benchmark Problems for Configuration}

We use the DASCMOP benchmark set~\cite{dascmop} to design the auto-MOEA/D. This set has nine constrained test functions: DASCMOP1-6, each with eleven constraints, and DASCMOP7-9, each with seven constraints. The constraints can be modified to consider three types of difficulties: type-I considers diversity-hardness, type-II considers feasibility-hardness and type-III considers convergence-hardness. More information about the problems and these difficulty triplets can be found in~\cite{dascmop}. We use the implementation of the test problems available from the \textit{Pymoo} python package~\cite{pymoo}. To have a more balancing training function set, we change the hardest constraints difficulty triplet, numbered 16,  from  our previous work~\cite{lavinas_gecco2022}, to a set of triplets ranging from low to high difficulty and numbered 4, 8, 12 and 16. 

\subsection{Benchmark Problems for Behavioral Analysis}

The real-world analytical multiobjective optimization problems used were selected from the new test suite introduced by Tanabe et. al \cite{TANABE2020}: (1) bar truss design (CRE21), the objectives are to minimize the structural weight and displacement resulting from the joint; (2) design of welded beams (CRE22), the objectives are to minimize the cost and the final deflection of the welded beams; (3) disc brake design (CRE23), the objectives are to minimize brake mass and decrease downtime; (4) side impact design of the car (CRE31), the objectives are to minimize the average weight of the car, the average speed of the column \textit{V} responsible for supporting the impact load and the force experienced by a passenger; (5) conceptual submarine project (CRE32), the objectives are to minimize the cost of transportation, the weight of the light vessel and the annual cargo transport capacity; (6) water resource planning (CRE51), the objectives are to minimize the cost of the drainage network, the cost of installing the storage center, the cost of installing the treatment center, the expected cost of flood damage and the expected economic loss due to the flood.

The real-world simulated MOPs used are the two recently proposed problems by the Japanese evolutionary computing society: (1) Mazda benchmark problem (MAZDA): this is a discrete optimizing problem to design MAZDA cars, where the objectives are to maximize the number of parts common to three different cars and minimize the total weight of the car~\cite{kohira2018proposal}; (2) lunar landing site selection (MOON): the goal is to select landing sites coordinates (x,y) for a lunar exploration robot to land with the objectives of minimizing the number of continuous shaded days, minimizing the inverse of the total communication time~\footnote{i.e. maximizing the total communication time.}, and minimizing the tilt angle~\cite{MoonOrbitingSatellite2015}.

\subsection{Evaluation Metrics for the Automatic Design}

Analysing MOP solvers considering only their final approximation provides limited information related to these algorithms' performance since any MOP solver should return a suitable set of solutions at any time during the search \cite{zilberstein1996using,radulescu2013automatically,anytimePLS,tanabe2017benchmarking}. Here, we analyse the anytime performance effects in terms of hypervolume (HV) values to investigate the impact of different configurations of MOEA/D on their Unbounded External Archive. We run auto-MOEA/D 10 times on each of the problems.

We use the following method to compare the results of the different strategies: we calculate the accumulative HV over the search progress to quantify the HV anytime performance. At every $1000$ evaluations, we calculate the HV of the solutions in the UEA at that iteration, using the reference point as: 11, over the number of objectives; following the work of Bezerra et al.~\cite{bezerra2016automatic}. Then, we sum all values to have an approximated evaluation of the anytime HV curve.


\begin{figure*}[htbp]
\centering
\includegraphics[width=1\textwidth]{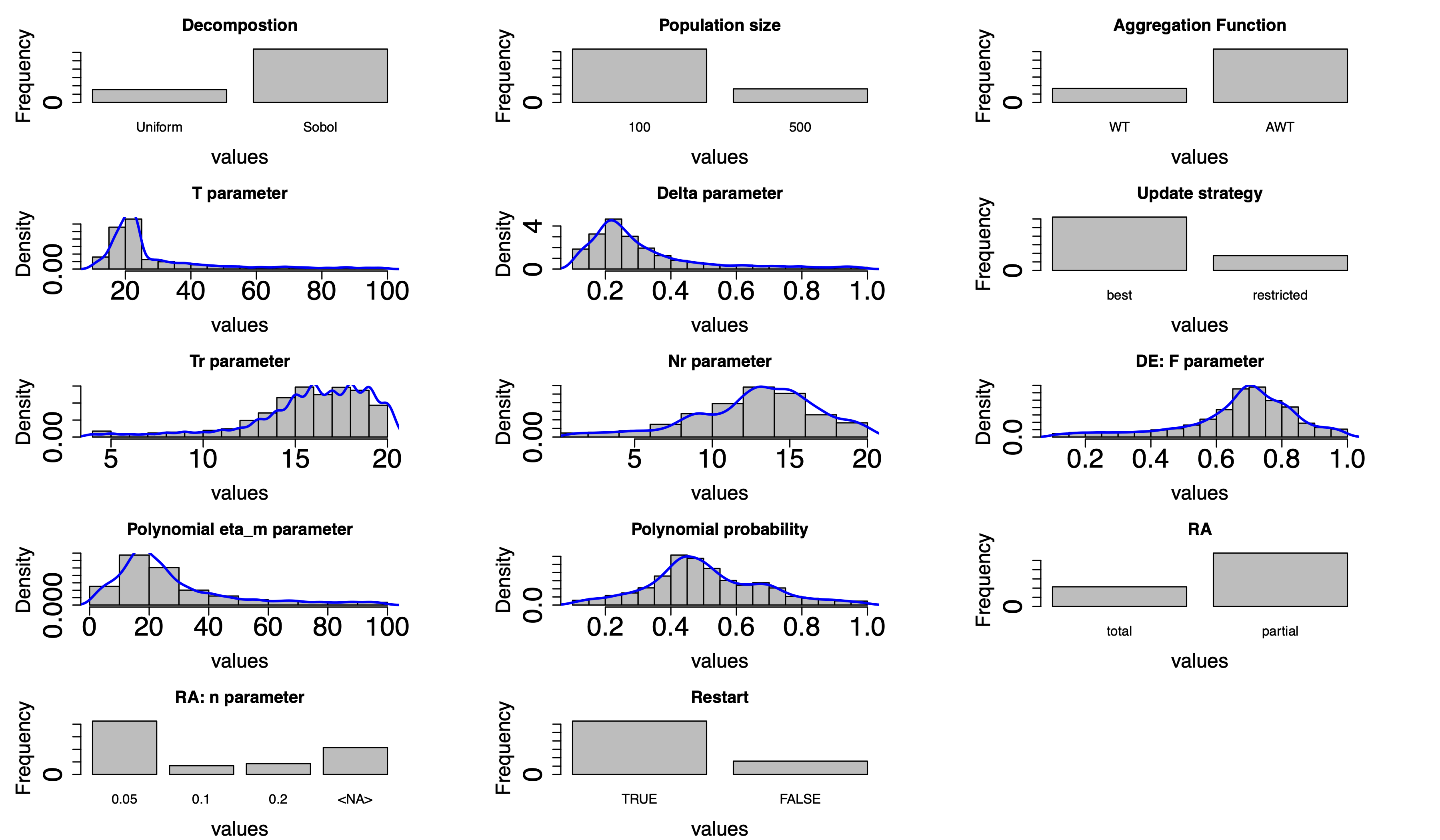}
\caption{irace output with the frequency of the different choices of components and parameters.}
\label{fig:stn_irace_output}
\end{figure*}

Figure~\ref{fig:stn_irace_output} shows the frequency of the different choices of components and parameters after the tuning and ablation design is performed. As in~\cite{LOPEZIBANEZ201643}, the ablation analysis is done to better use the available computational resources, focusing on better-performing configurations found by an automatic configuration method, and proceeds by changing one parameter at a time, focusing more on well-performing configuration, instead of more common useful configurations.

There is a consensus over the components and parameters studied here. This suggests that irace is confident that the choice of components used in the auto-MOEA/D design has at least an adequate overall performance in the problems used during the configuration process.

\subsection{Performance of auto-MOEA/D}\label{trends_auto}

Here, we briefly describe the performance of auto-MOEA/D in terms of HV~\footnote{higher is better.} and variance of the final population, and the STNs metrics: number of nodes, edges and shared nodes. For the calculation of HV, the objective function was scaled to the (0, 1) interval, with the reference point of $1.1$, repeated over the number of objectives. Thus, the maximum HV is $1.21$ for MOPs with two objectives or $1.331$ for MOPs with three objectives after scaling the values. Instead of showing the achieved HV value of the auto-MOEA/D, we calculate how close the performance of this algorithm is to the maximum possible HV value (max(HV)). Values close to 0 indicate no HV performance~\footnote{probably the approximation found by auto-MOEA/D is out of the range of the reference point.} while values close to 1 indicate high performance.

\begin{table}[htbp]
\centering
\caption{hypervolume ratio \textit{(HV)/max(HV)}, HV \textit{S}tandard \textit{D}eviation,  Standard \textit{Deviation},  \textit{Nodes}, \textit{Edges}, population \textit{variance}, and the number of solutions in the theoretical Pareto front (\textbf{$\#PF$}) of the auto-MOEA/D for the DASCMOP problems.}
\label{metrics_ablation}
\begin{tabular}{c|cccccc}
    \rowcolor[gray]{.75}  & \multicolumn{6}{c}{\textbf{auto-MOEA/D}} \\
    \rowcolor[gray]{.82}MOP & HV/max(HV) & SD & Nodes &  Edges & Variance & $\#PF$ \\ 
    DASCMOP1 & 0.931 & 0.002 & 4268 & 4812 & 0.514 & 212 \\
    DASCMOP2 & 0.964 & 0.002  & 4821 & 5480 & 0.45 & 279 \\
    DASCMOP3 & 0.962 & 0.007 & 7175 & 8595 & 0.494 & 44 \\
    DASCMOP4 & 0 & 0 & 2316 & 2890 & 4.812 & 0 \\
    DASCMOP5 & 0.095 & 0.362 & 2352 & 2998 & 5.089 & 0 \\
    DASCMOP6 & 0 & 0 & 2313 & 2901 & 5.022 & 0 \\
    DASCMOP7 & 0.035 & 0 & 2388 & 3139 & 5.035 & 0 \\
    DASCMOP8 & 0 & 0 & 2465 & 3235 & 4.372 & 0 \\
    DASCMOP9 & 0.849 & 0.397 & 15756 & 19233 & 0.437 & 10 \\
    \rowcolor[gray]{.95}MOON & 0.392 & 0.452 & 1375 & 3012 & 0.007 & 2 \\
    \rowcolor[gray]{.95}MAZDA & 0.016 & 0.026 & 5443 & 5531 & 5.366 & 0 \\
    CRE21 & 0.989 & 0.011 & 4774 & 6370 & 0.1 & 0 \\
    CRE22 & 0.995 & 0.001 & 2047 & 2192 & 0.119 & 12 \\
    CRE23 & 0.608 & 0.003 & 7774 & 10369 & 0.335 & 1 \\
    CRE31 & 0.567 & 0.002 & 5259 & 5497 & 0.72 & 2 \\
    CRE32 & 0.756 & 0.011 & 14976 & 17743 & 0.331 & 0 \\
            
\end{tabular}
\end{table}

The different metric values for the auto-MOEA/D are shown in Table~\ref{metrics_ablation}. We can see that higher HV values correspond to a high number of nodes and edges and lower final populational variance values for the DASCMOP1, DASCMOP2, DASCMOP3 and DASCMOP9 problems. In contrast, the opposite happens for the other DASCMOP problems. Given the low HV performance, we speculate that auto-MOEA/D has a premature convergence probably to local optima areas of the decision space. 

The HV performance of auto-MOEA/D deteriorates for the simulated MOON and MAZDA problems. There is no agreement in values of the number of nodes, edges and variance found for auto-MOEA/D in this set of problems, suggesting that each problem has different features in relation to the other. Interestingly, the number of nodes and variance metrics are the lowest in the MOON problem which might suggest that the population gets trapped in local optima during the run. This could also explain partially why the HV performance is low. For the MAZDA problem auto-MOEA/D also has a poor HV performance, but a higher number of nodes, edges and variance in comparison to the MOON problem.

For the CRE problems,  we see that the HV performance is close to the DASCMOP1, DASCMOP2, DASCMOP3 and DASCMOP9 problems. In general, the other metrics follow a similar comparison trend between CRE problems and DASCMOP. Thus, we understand that there are similarities among the artificial DASCMOP problems and the CRE problems and limited similarities with the more challenging simulated MOON and MAZDA problems.

Finally, we comment about the results on constraint difficulty change between our previous work and this current study. On contrary to our expectations, having different constraints difficulty levels lead to improvements the HV performance of auto-MOEA/D in the easy problems (DASCMOP1-3 and 9), but no clear increments in performance for the hard set of problems (DASCMOP4-8).

\section{Comparison of the Components}\label{search_behavior}

Here we use our methodology to investigate the effects of the final configuration of a machine-designed multiobjective algorithm. This analysis aims to measure the differences in the decision and objective space dynamics among several variants from the MOEA and, through these measures, identify the most influential components of the automatically designed algorithm.

\begin{table}[htbp]
\caption{Auto-MOEA/D setup and the variants under analysis. For each variant, only \textit{one component} is changed, while the other components are the same as auto-MOEA/D.}
\label{variants}
\begin{tabular}{c|c}
\rowcolor[gray]{.75}\multicolumn{2}{c}{\textbf{Auto-MOEA/D setup, and its Variants}}\\ 
\rowcolor[gray]{.85}\textbf{auto-MOEA/D}                    & \textbf{Component variant} \\ 

\textbf{Decomposition + pop. size}      &   \textbf{Decomposition + pop. size} \\       $Sobol, 100$                   & $SLD, 300$                    \\ \hline
\textbf{Aggregation function}      &   \textbf{Aggregation function} \\           $AWT$    & $WT$                       \\ \hline
\textbf{Update}      &   \textbf{Update} \\
$Best$    & $Restricted$           \\ 
$nr = 9$        & $nr = 2$         \\  \hline

\textbf{Neighbourhood pars.}      &   \textbf{Neighbourhood pars.} \\
$T = 22$    &  $T = 20$                   \\
$Delta = 0.9822$    &   $Delta = 0.9$       \\ \hline

\textbf{Operators pars.}      &   \textbf{Operators pars.} \\
$DE: F = 0.4908$    &  $DE: F = 0.5$                          \\
Polynomial: $\eta_m = 80.9844$    &  Polynomial:  $\eta_m = 20$     \\
Polynomial: $prob = 0.4556$    &   Polynomial: $prob =  0.3$   \\ \hline

\textbf{Restart}      &   \textbf{Restart} \\                           $True$    & $False$                           \\  \hline
\textbf{RA}      &   \textbf{RA} \\                           $Partial, 5\%$    & $False$                           \\ 

\end{tabular}
\end{table}


To analyse these behavioral effects of the different variants, we compare the auto-MOEA/D described in Section~\ref{exp_automatic} against variants with at most one single component altered. We obtain these variants by changing or removing a single component of the auto-MOEA/D at a time. 
This is done by either (1) removing the component from the algorithm when possible or (2) changing its parameters to its counterpart in the traditional MOEA/D. 

Table 4 lists the auto-MOEA/D on the left side, which is generated by
the algorithm in section 4, and the variants on the right side. These
variants are not necessarily components generated by auto-MOEA/D. When total removal of the component is not possible, we use the standard version of these components introduced by Hui Li and Qingfu Zhang in MOEA/D-DE~\cite{moeadde}, and commonly found in the literature. Thus; there are \textit{seven} variants to analyse, which added to auto-MOEA/D itself least to a total of \textit{eight} algorithmic variations of the MOEA/D framework. These variations are compared quantitatively and visually in terms of their STN models to detect which components procude the most extensive changes.

\section{behavioral dynamics}\label{dynamics}

To quantitatively analyse the dynamics of the search progress of the different variants of MOEA/D, we model the search dynamics using STNs for each pair of auto-MOEA/D and an auto-MOEA/D variant, leading to seven different pairs. We highlight that given the lower impact on performance of the decomposition and population size observed on the original paper (pre-extension, see~\cite{lavinas_gecco2022}) and since that most works on the literature focus on population size and decomposition methods(to name only a few~\cite{behavior_popsize_continuous,lavinas2022faster,metrics_resolution_aguirre,app_resolution_aguirre,bezerra2015comparing}), we combine these components into one, and direct our studies towards the other components.

We base our quantitative analysis on the traditional multiobjective metrics hypervolume (HV); the number of: nodes, edges and  shared nodes of the STN models; and the populational variance. For the HV, we use the reference point of 1.1, repeated over the number of objectives. We linearly scaled all objectives to the interval $[0, 1]$, for a more straightforward comparison among the algorithms. We run each variant 10 times on each of the problems.

\begin{figure*}[htbp]
\centering
	\begin{subfigure}[!t]{0.3\textwidth}
    	\includegraphics[width=1\textwidth]{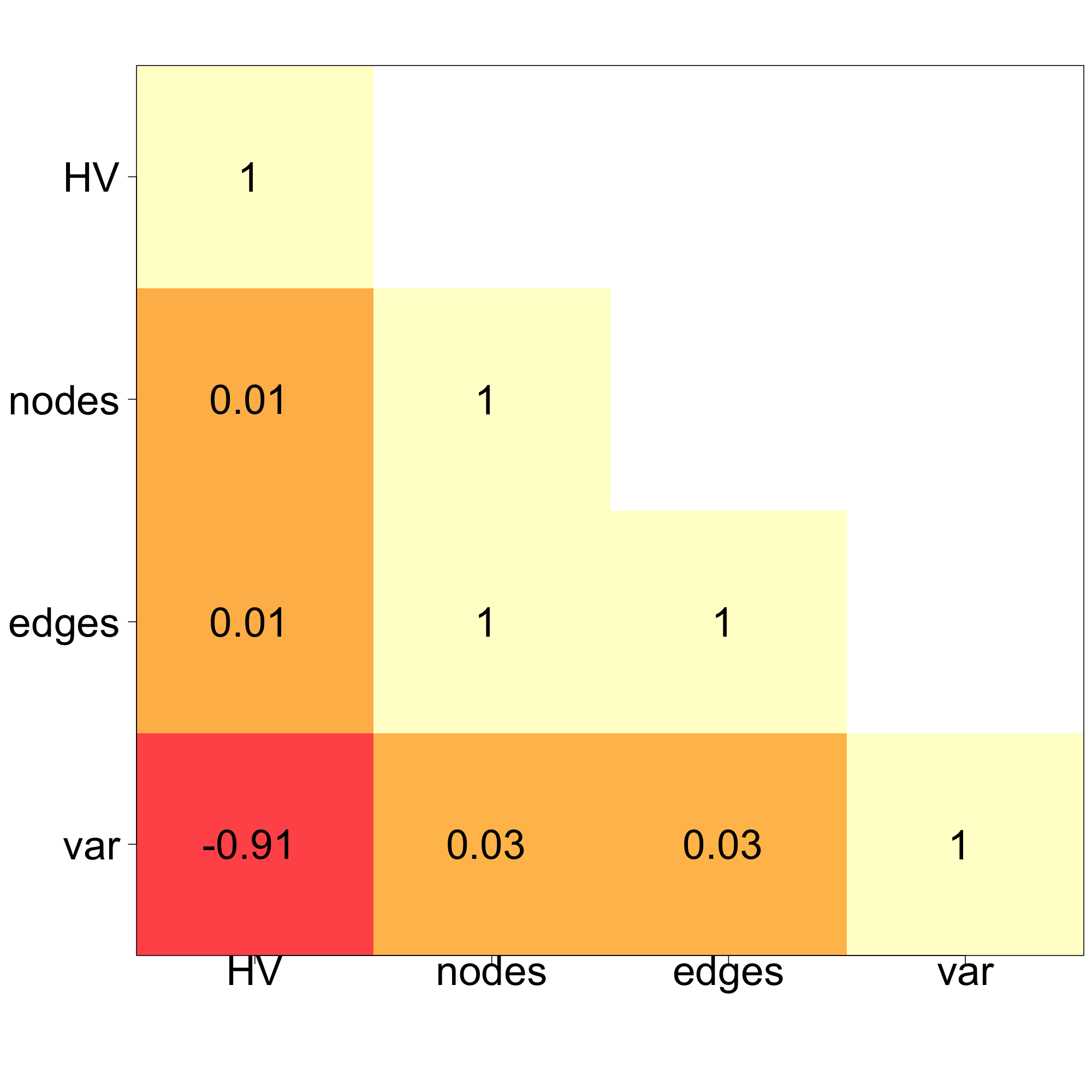}
        \caption{DASCMOP problems.}
	\end{subfigure}
    \begin{subfigure}[!t]{0.3\textwidth}
    	\includegraphics[width=1\textwidth]{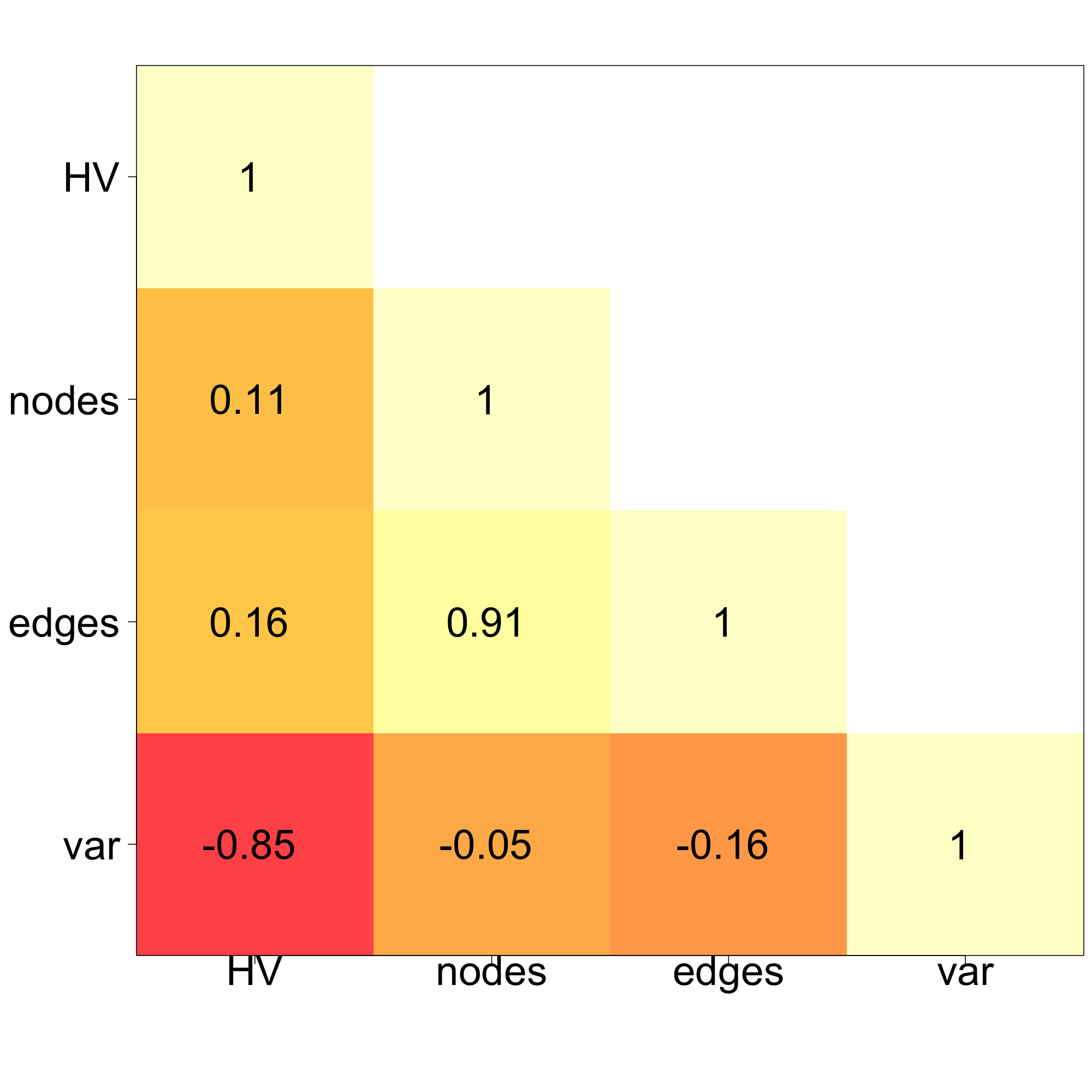}
        \caption{Simulated problems.}
	\end{subfigure}
    \begin{subfigure}[!t]{0.3\textwidth}
    	\includegraphics[width=1\textwidth]{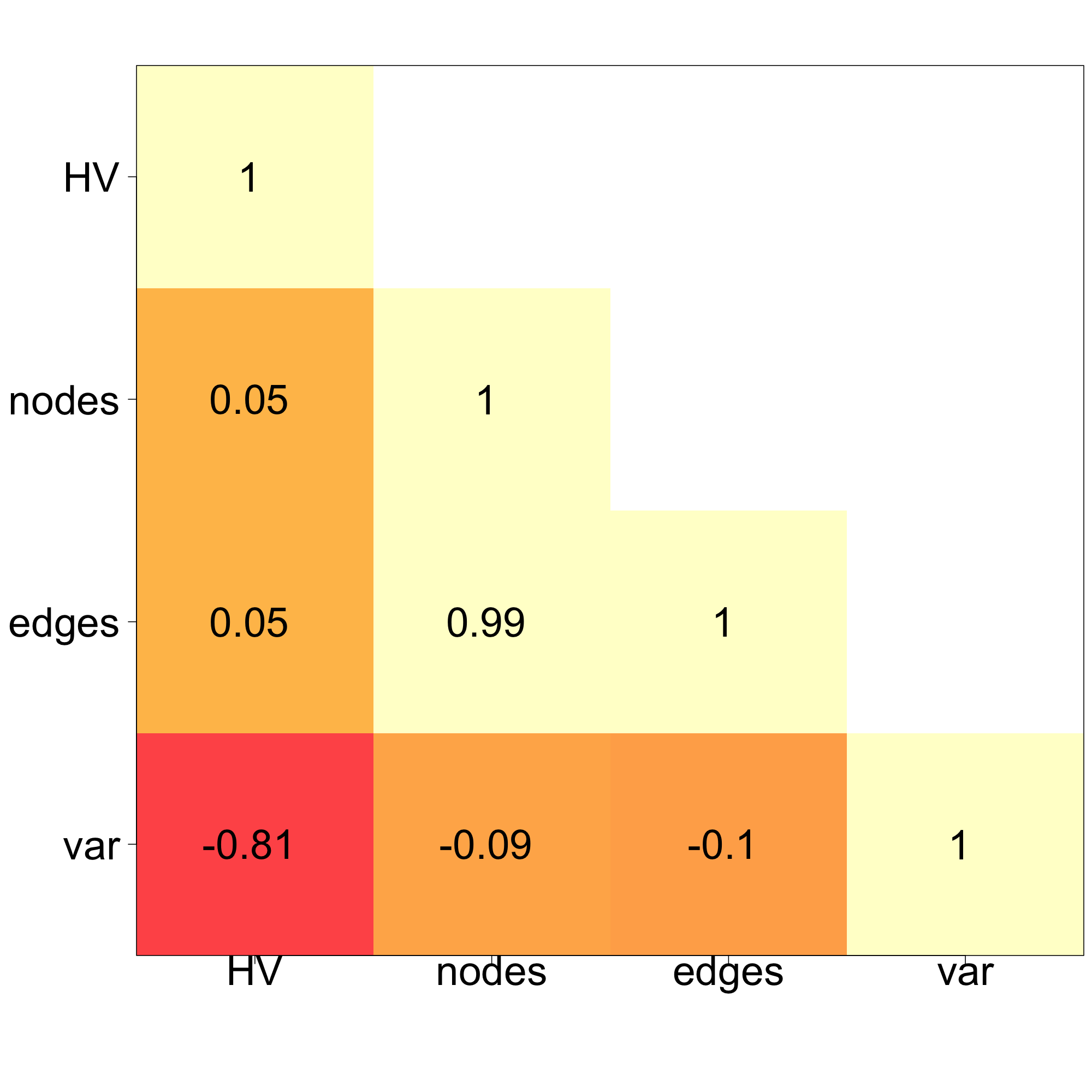}
        \caption{Analytical problems.}
	\end{subfigure}
    \caption{Correlation matrix among the different metrics. We find some correlation among HV and \textbf{var}iance of the final population; and correlation between the number of nodes and edges.}
    \label{fig:correlation_all}
\end{figure*}

\subsection{Metrics Analysis}

Figure~\ref{fig:correlation_all} shows the correlation matrix among the different metrics studied, considering the results of the MOEA/D variants for the DASCMOP, simulated~\footnote{MOON and MAZDA.} and analytical~\footnote{CRE family.} problems. Since the number of nodes and edges have a high correlation, we remove the number of edges metric from our analysis. The other STN metrics, together with the final population variance are on the decision space analysis; thus, we use them in our following analysis to strengthen the study. Furthermore, we see a correlation between the HV metric and the population variance. This correlation suggests a link between this decision space metrics and HV increments of performance. Moreover, we understand that there might exist a connection between solution diversity, represented by the variance, and overall performance and that this connection is problem-independent.

\subsection{Objective Space Behavior}

\begin{figure*}[htbp]
    \centering
	\includegraphics[width=1\textwidth]{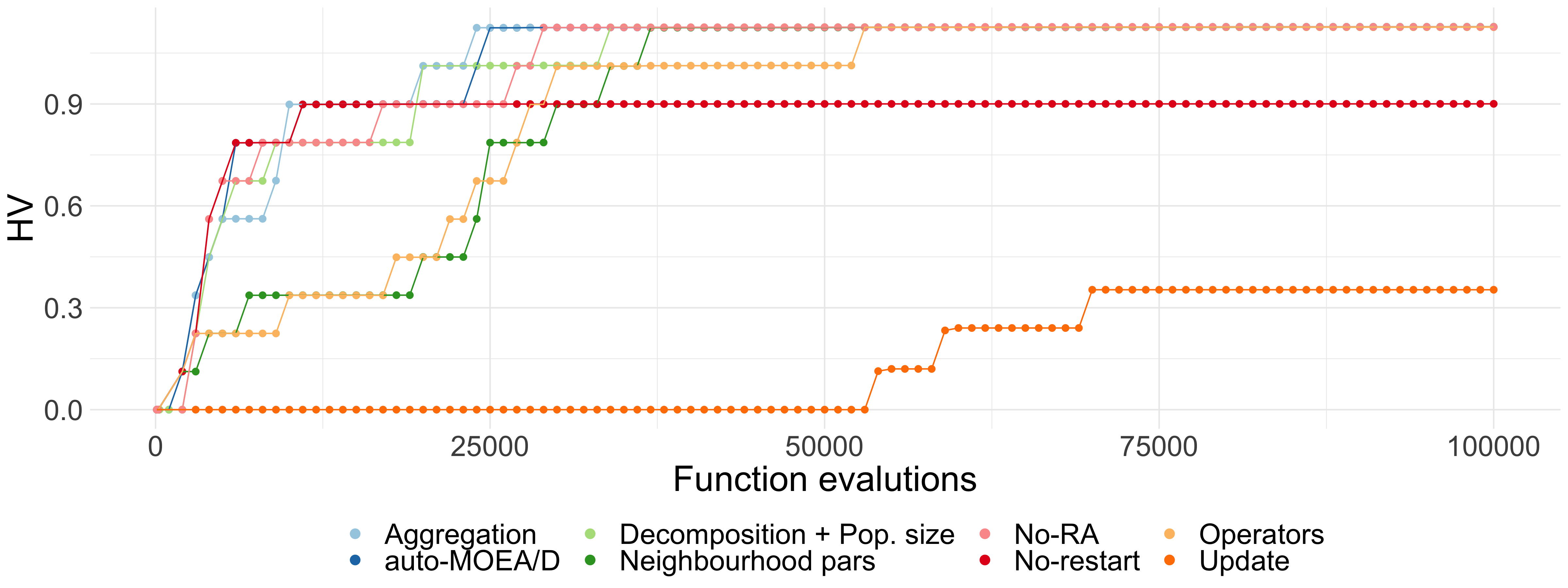} 
	\caption{DASCMOP1 - HV anytime performance of auto-MOEA/D and its variants. The update strategy and not using restart lead to worse performance.}
\end{figure*}

\begin{figure*}[htbp]
    \centering
	\includegraphics[width=1\textwidth]{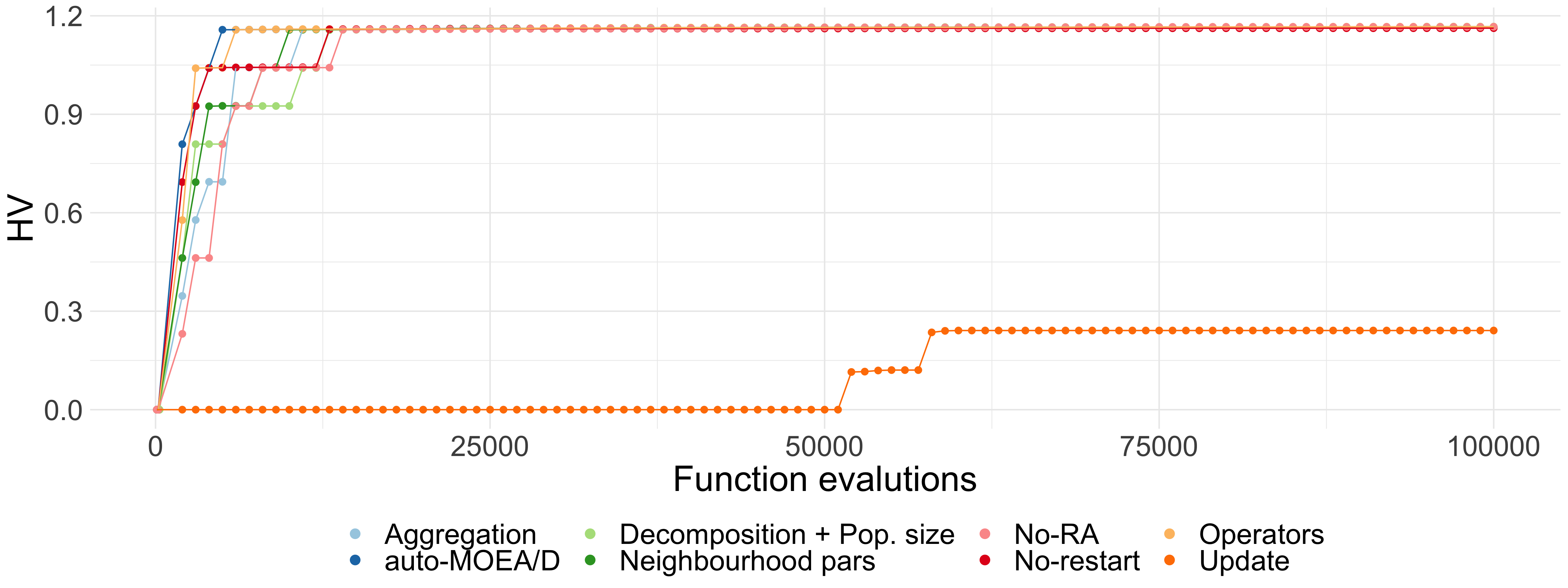} 
	\caption{DASCMOP2 - all variants converge to the maximum HV, with the exception of the update strategy.}
\end{figure*}

\begin{figure*}[htbp]
    \centering
	\includegraphics[width=1\textwidth]{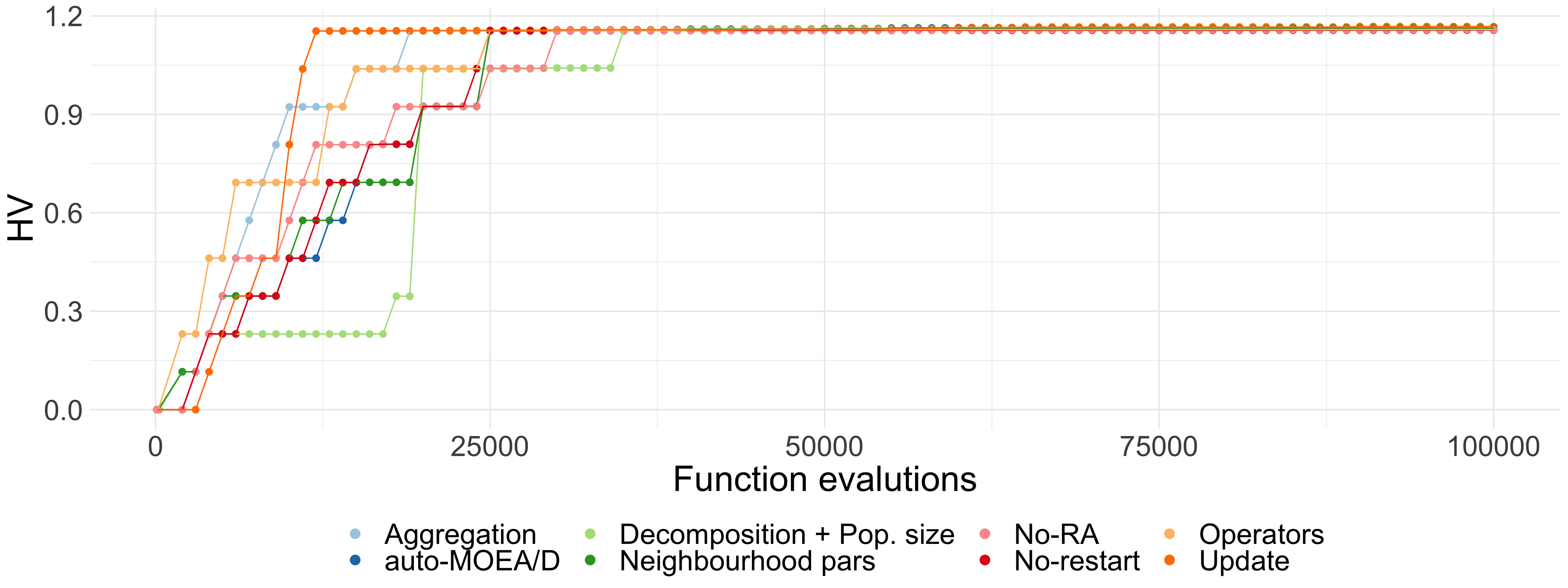} 
	\caption{DASCMOP3 - all variants converge to the maximum HV before half of the search.}
\end{figure*}

\begin{figure*}[htbp]
    \centering
	\includegraphics[width=1\textwidth]{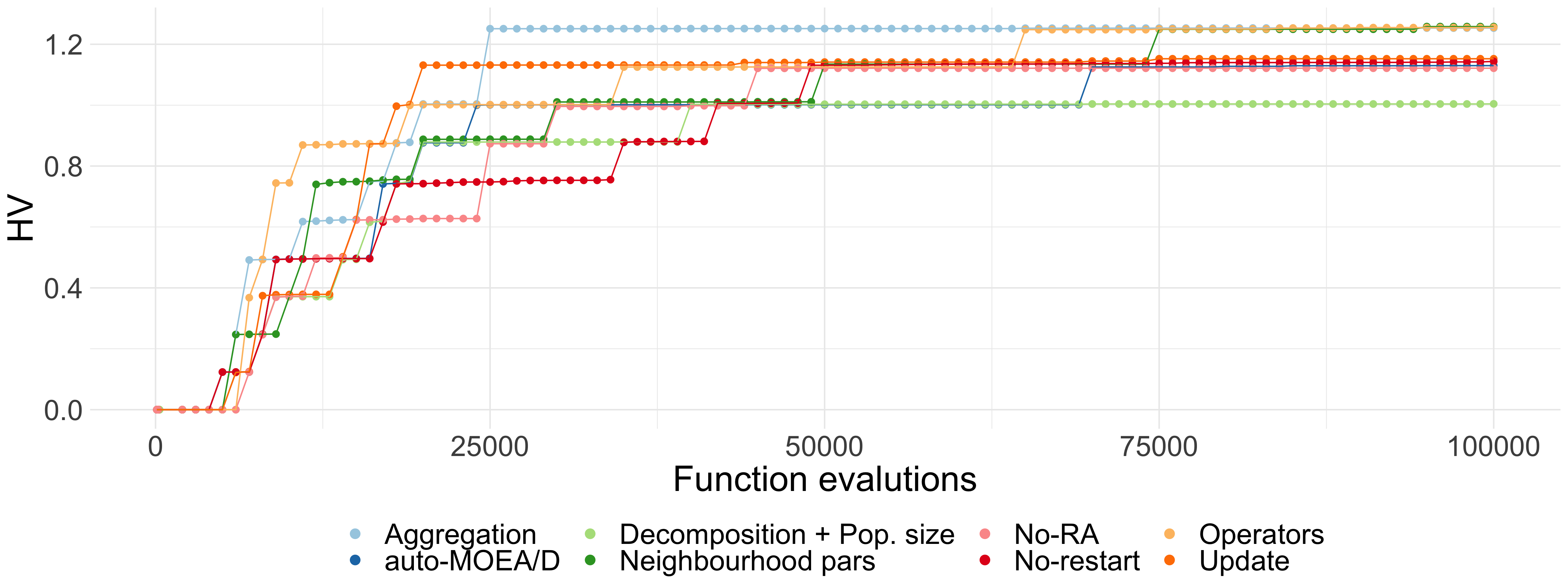} 
	\caption{DASCMOP9 - the aggregation variant converges to high HV values faster than the other variants.}
	\label{fig:anytime_Dascmop}
\end{figure*}

\begin{figure*}[htbp]
    \centering
	\includegraphics[width=1\textwidth]{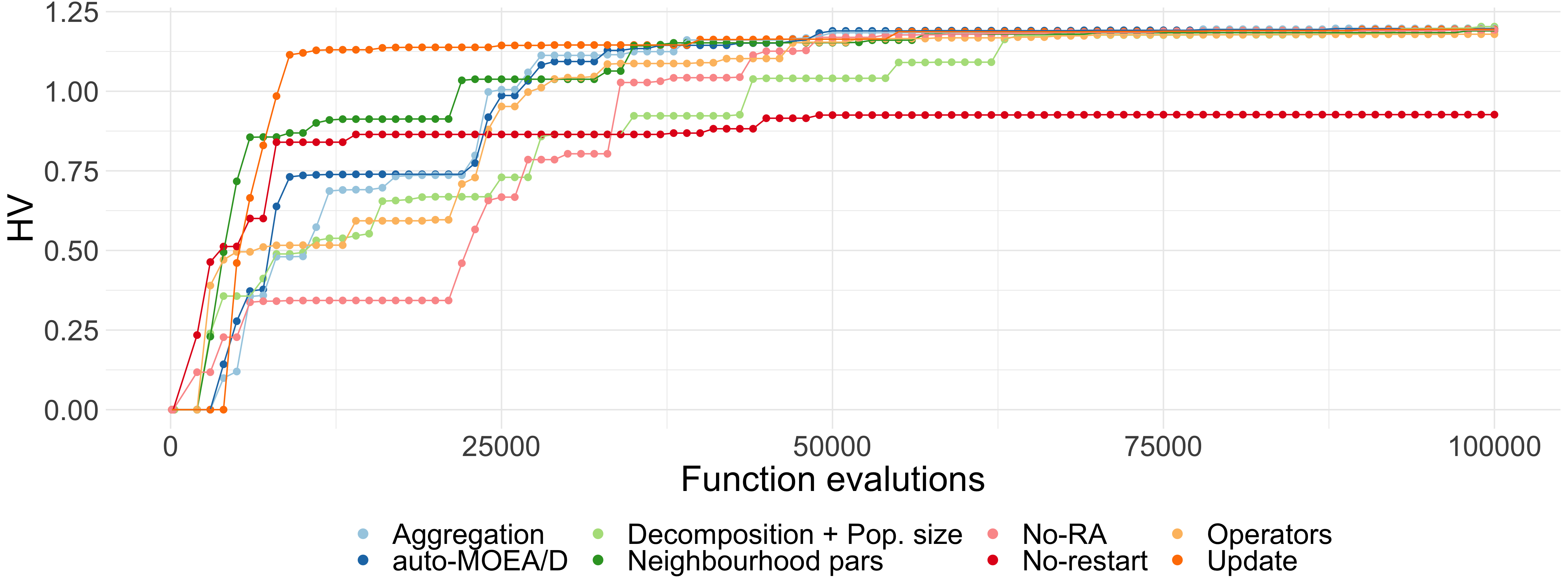}
	\caption{CRE21 - the variants converge to about the same HV value, with the exception of the no-restart.}
	\label{fig:anytime_CRE21}
\end{figure*}

\begin{figure*}[htbp]
    \centering
	\includegraphics[width=1\textwidth]{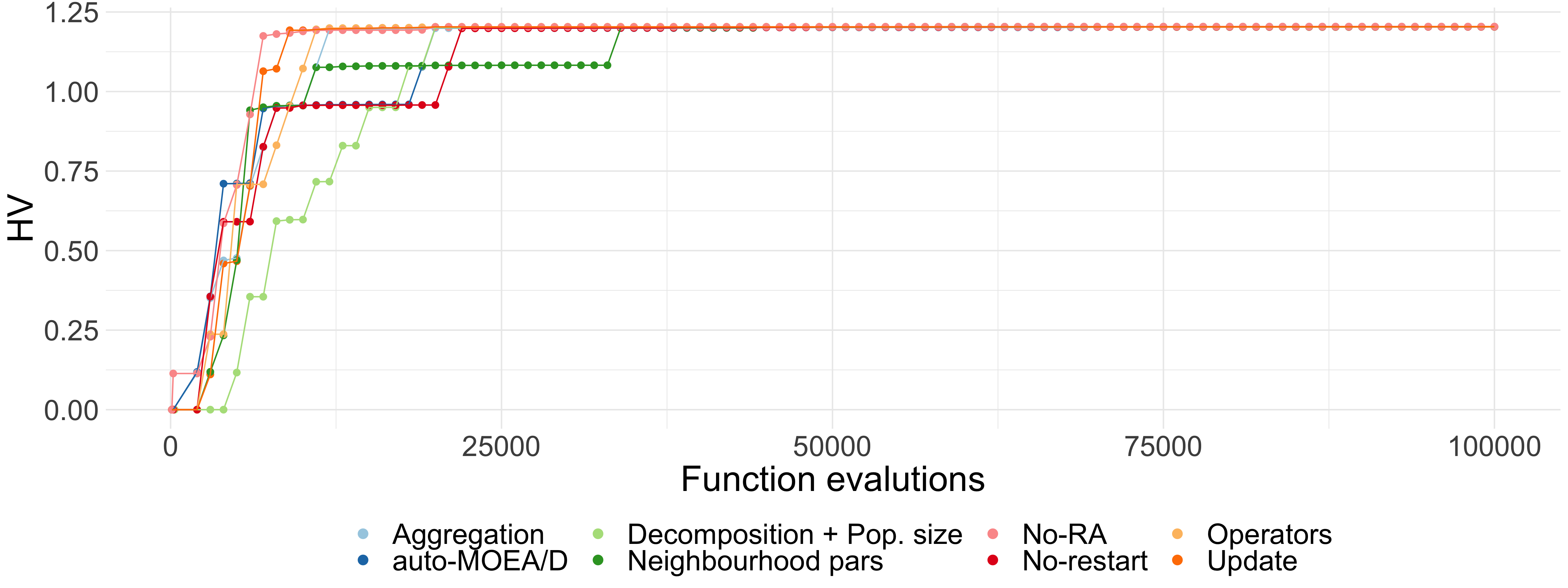}
	\caption{CRE22 - all variants converge to maximum HV at about one third of the search.}
	\label{fig:anytime_CRE22}
\end{figure*}

\begin{figure*}[htbp]
\centering
	\includegraphics[width=1\textwidth]{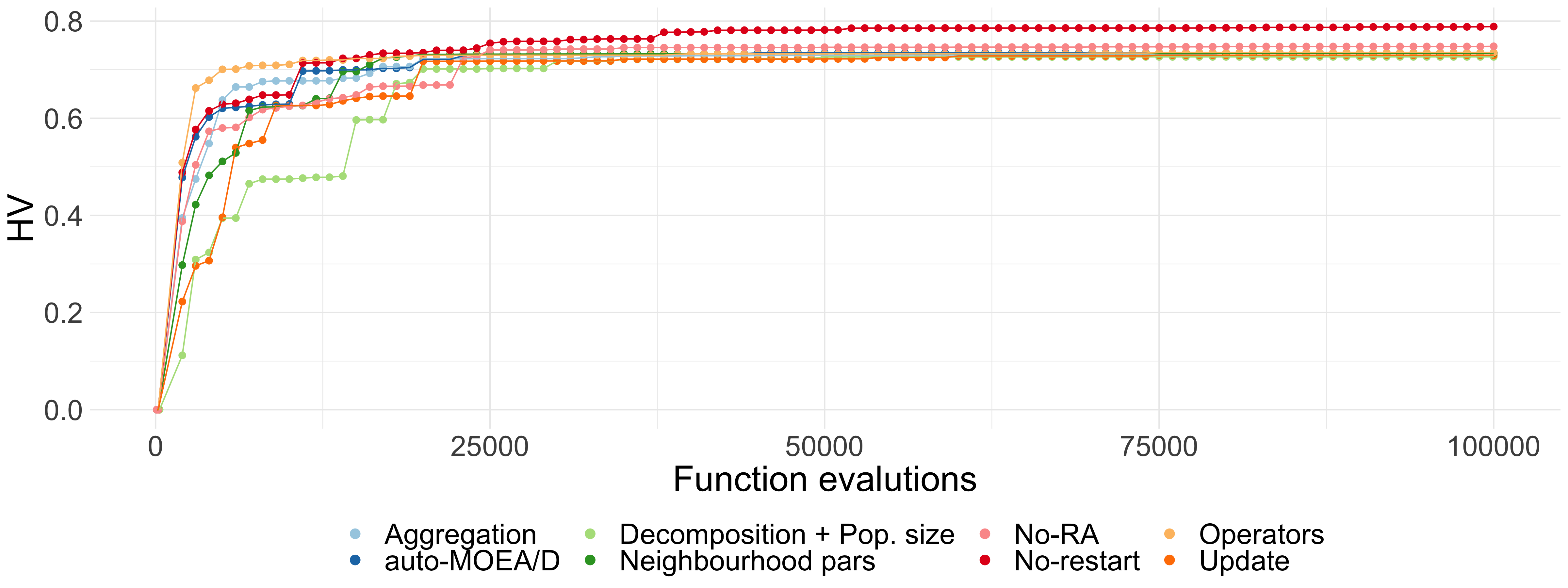} 
	\caption{CRE23 - the variants converge to sub-optimal HV values at about one third of the search.}
	\label{fig:anytime_CRE23}
\end{figure*}

\begin{figure*}[htbp]
    \centering
   	\includegraphics[width=1\textwidth]{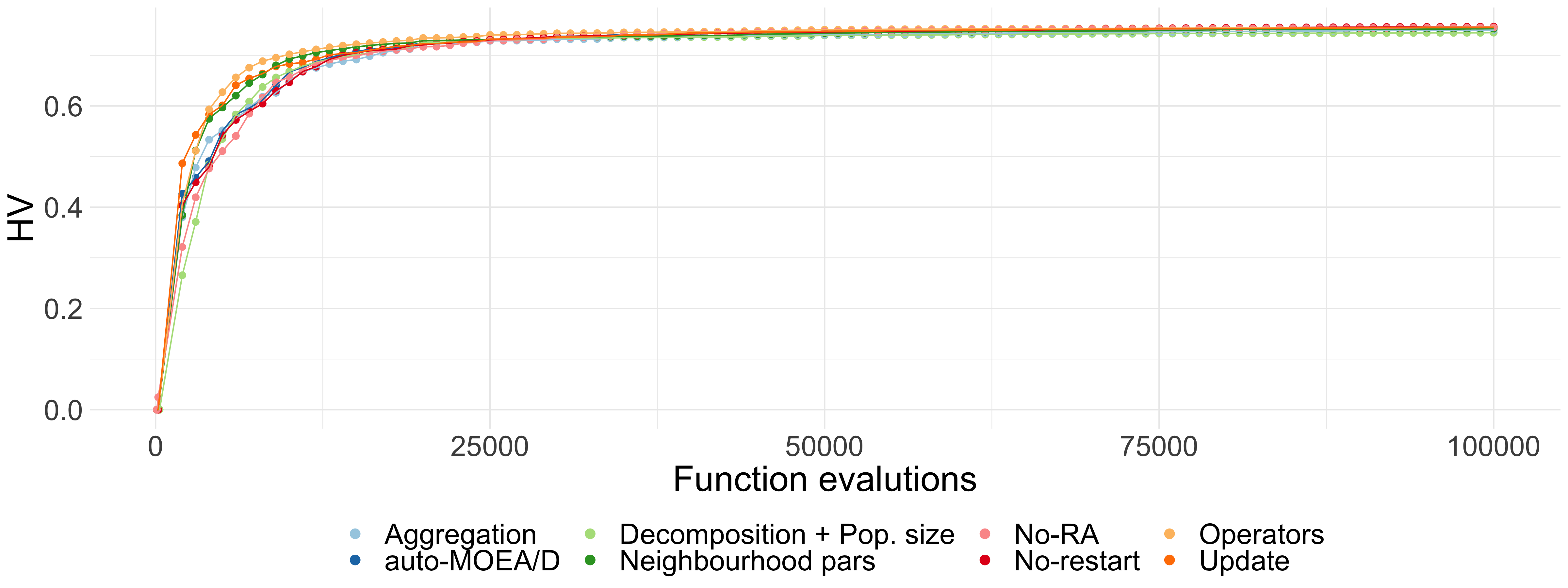}
	\caption{CRE31 - similarly for CRE23, the variants converge to sub-optimal HV values at about one third of the search.}
	\label{fig:anytime_CRE31}
\end{figure*}

\begin{figure*}[htbp]
    \centering
    \includegraphics[width=1\textwidth]{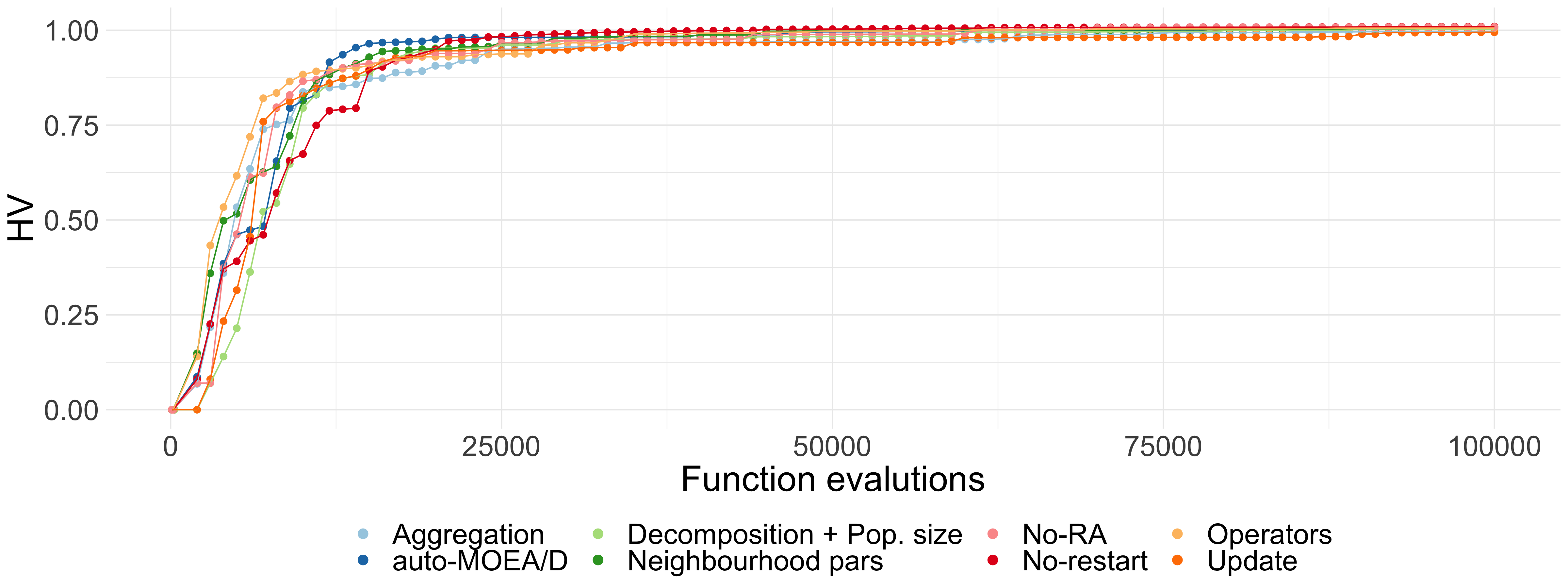}
    \caption{CRE32 - Better performance than CRE31, but the variants again converge early to sub-optimal values.}
    \label{fig:anytime_CRE32}
\end{figure*}

\begin{figure*}[htbp]
    \includegraphics[width=1\textwidth]{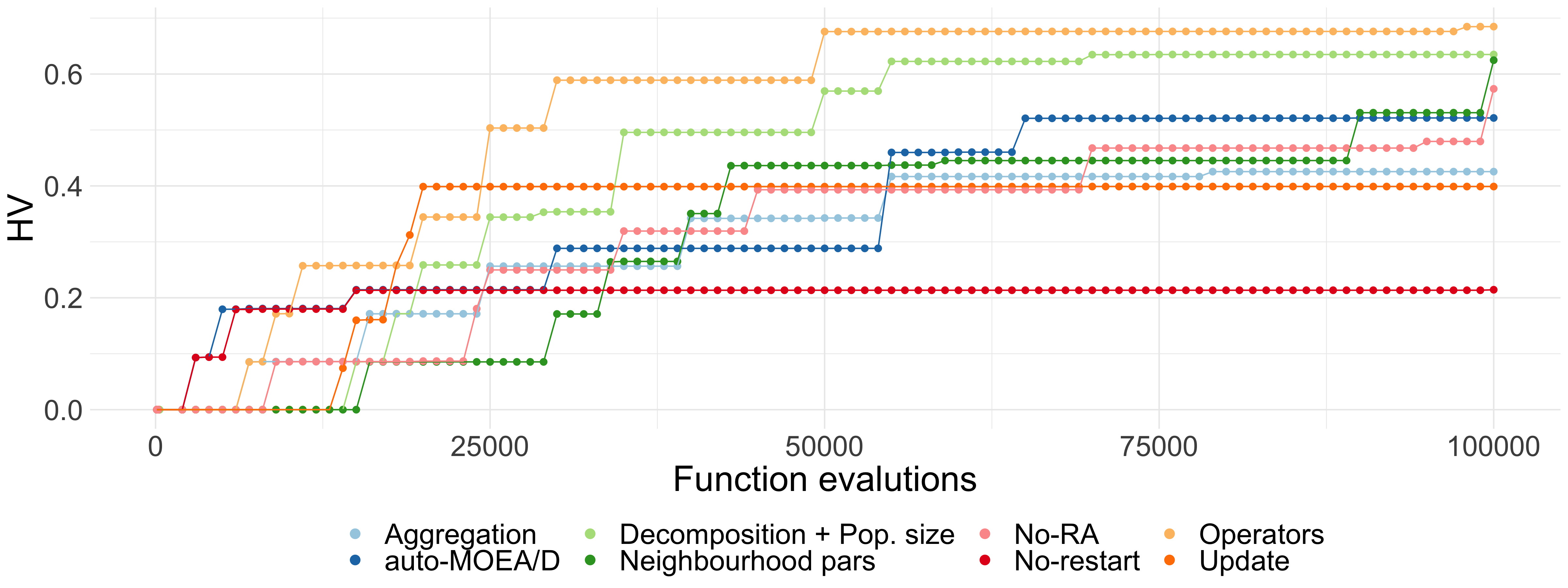}
	\caption{MOON - the HV values are much different among the variants, with the operators variant achieving the best results and not-restarting has a big negative impact.}
	\label{fig:anytime_MOON}
\end{figure*}

\begin{figure*}[htbp]
    \centering
	\includegraphics[width=1\textwidth]{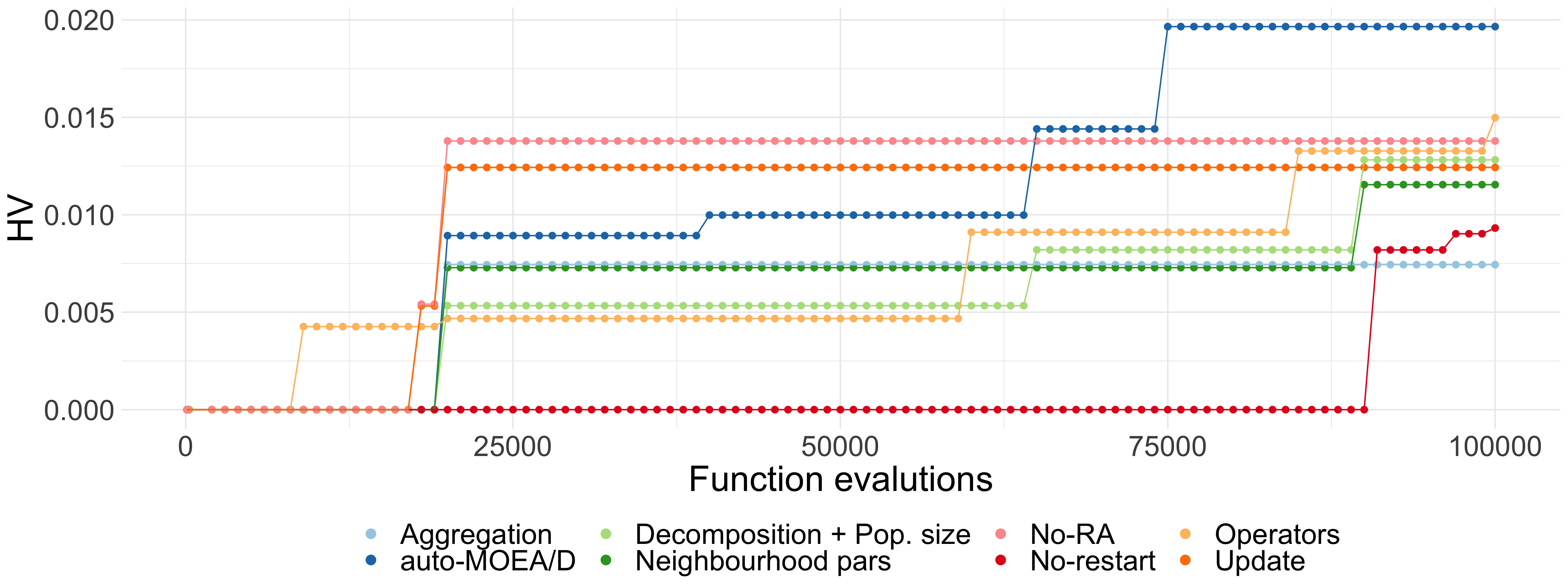} 
	\caption{MAZDA - all variants perform badly, with auto-MOEA/D achieving the highest HV value.}
	\label{fig:anytime_MAZDA}
\end{figure*}


Here, we analyse the anytime performance effects in terms of HV values to investigate the impact of different components variants and auto-MOEA/D in analytical and simulated real-world problems. We first start our analysis using the DASCMOP1-3 and 9, Figure~\ref{fig:anytime_Dascmop}, as auto-MOEA/D has a very poor performance in the other problems~\footnote{All Figures for the anytime HV performance are available in Zenodo \href{https://zenodo.org/XXXXX/}{https://zenodo.org/XXXXX/}}. We can see that for these problems, increments in HV values are followed by periods without changes in performance for almost all of the variants. We can also see that the variants converge before half of the search, except for the update variant in the DASCMOP1-2 problems. For DASCMOP9, the convergence happens a little later, but the HV curves for this problem are overall similar to the other problems.

For the CRE problems, all auto-MOEA/D variants converge at around 25000 evaluations, as we can see in Figures~\ref{fig:anytime_CRE21},~\ref{fig:anytime_CRE22},~\ref{fig:anytime_CRE23},~\ref{fig:anytime_CRE31} and \ref{fig:anytime_CRE32}. The exception is for the CRE21 problem, in Figure~\ref{fig:anytime_CRE21}, where not using restart converged to a lower HV value than the other variants. This is similar to the convergence behavior we found for the DASCMOP problems, although the curves here are more balanced. It seems that DASCMOP1-3 and 9 and the CRE problems have little impact on the ability of the auto-MOEA/D variant to perform well in terms of anytime HV performance.

The same observation cannot be made for the MAZDA and MOON problems. We can see in Figures~\ref{fig:anytime_MOON} and \ref{fig:anytime_MAZDA} that most of the variants have trouble improving the HV over the evaluations and that the best variant is problem-dependent. For both problems, increments in HV values are followed by periods without changes in performance for almost all of the variants. This is similar behavior to the DASCMOP problems; however, here we see that the periods without increments are much longer. Interestingly, for the MAZDA problem  not-using restart has generally very low performance during almost all of the search progress to only at the very end improving the performance substantially. 

\subsection{Decision Behavior Dynamics}

Based on the results shown above, we compared the auto-MOEA/D and its variants in terms of HV, final population variance, and the STNs metrics: number of nodes, shared nodes and the number of solutions in the best approximation to the Pareto front~\footnote{for the DASCMOP this is the theoretical Pareto front}.. For a more straightforward comparative analysis, we calculate the difference between the results found by the variants to the results found by the auto-MOEA/D (Table~\ref{metrics_ablation}). We show the $\Delta$HV, $\Delta$nodes, $\Delta$variance~\footnote{metric value of auto-MOEA/D minus metric value of the variant.}. For all of these metrics, positive values indicate larger values in relation to the base algorithm, while negative values indicate the opposite. The number of shared nodes and solutions in the Pareto front are the two absolute metrics.

\begin{table*}[htbp]
\vspace{-1em}
\centering
\caption{\textit{$\Delta$HV}; number of \textit{$\Delta$nodes}; \textit{$\Delta$variance} of the population, number of \textit{Shared} nodes and the number of solutions in the Pareto front (\#PF) given the different variants of the auto-MOEA/D in terms of components on the constrained problems.}
\vspace{-1em}
\label{metrics_all}
\footnotesize
\begin{tabular}{c|cccccc|cccccc}
    
    \rowcolor[gray]{.75}  &\multicolumn{6}{c|}{\textbf{Decomposition + pop. size variant}} & \multicolumn{6}{c}{\textbf{Aggregation function variant}}\\
    \rowcolor[gray]{.82} MOP & $\Delta$HV & SD &  $\Delta$nodes  & $\Delta$variance & Shared & \#PF & $\Delta$HV & SD &  $\Delta$nodes  & $\Delta$variance  & Shared & \#PF \\ 
    DASCMOP1  &  -0.001 & 0 & 313  &  -0.012  &  1  &  478 
    &  0  & 0.002 & -86  &  0.03  &  703  &  424 \\ 
    DASCMOP2  &  -0.001  & 0 & 757  &  -0.005  &  3  &  556 
    &  0  & 0.002 & 90  &  0.007  &  743  &  611 \\ 
    DASCMOP3  & 0 & 0  & 1884  &  0.026  &  3  &  66 
    &  0.003  & 0.007 & 59  &  0.077  &  790  &  78 \\ 
    DASCMOP4  &  -0.116 & 0.043 & 103  &  -1.847  &  4  &  0 
    &  0  & 0 & -10  &  -0.291  &  2141  &  0 \\ 
    DASCMOP5  &  0  & 0.058 & 183  &  -1.532  &  5  &  0 
    &  0.114  & 0.041 & -6  &  -0.13  &  2168  &  0 \\ 
    DASCMOP6  &  -0.113 & 0.041 & 159  &  -1.138  &  5  &  0 
    &  0 & 0 & 0  &  -0.112  &  2140  &  0 \\ 
    DASCMOP7  &  0 & 0 & 178  &  -1.315  &  7  &  0 
    &  0  & 0 & 26  &  0.015  &  2196  &  0 \\ 
    DASCMOP8  &  -0.124  & 0.391 & 234  &  -0.835  &  7  &  0 
    &  0 & 0 & 15  &  -0.108  &  2262  &  0 \\ 
    DASCMOP9  &  0.126  & 0.53 & 3483  &  0.006  &  6  &  12 
    &  -0.123  & 0.024 & 1162  &  -0.011  &  2618  &  17 \\ 
   
    \rowcolor[gray]{.95}MOON  &  -0.113 & 0.371 &  272  &  -0.076  &  343  &  2 
      &  0.096 & 0.451 & 7  &  0.004  &  911  &  2 \\  
    \rowcolor[gray]{.95}MAZDA  &  0.007 & 0.02 & 3682  &  -4.762  &  5  &  0 
      &  0.012  & 0.023 & -118  &  0.144  &  1790  &  0 \\  
   
    CRE21  &  -0.007 & 0.007 & 1753  &  0  &  192  &  0   
      &  -0.001 & 0.008 & 22  &  0.034  &  4562  &  0 \\  
    CRE22  &  0 & 0.001 & 198  &  0.064  &  27  &  19   
      &  0  & 0.001 & 69  &  -0.033  &  551  &  15 \\  
    CRE23  &  0.009  & 0.004 & 2862  &  -0.007  &  374  &  3 
      &  0.001 & 0.005 & -101  &  -0.014  &  1362  &  2 \\  
    CRE31  &  0.01  & 0.001 & 1433  &  -0.01  &  4  &  5 
      &  0.004  & 0.004 & 439  &  0.048  &  556  &  7 \\  
    CRE32  &  0.003 & 0.002 & 6693  &  0.025  &  17  &  0 
      &  0.01 & 0.019 & 339  &  0.109  &  1267  &  0 \\

    \rowcolor[gray]{.75} &\multicolumn{6}{c|}{\textbf{Update variant}} & \multicolumn{6}{c}{\textbf{Neighbourhood parameter variant}}\\
   
    DASCMOP1  &  0.774 & 0.569 & 954  &  -0.145  &  258  &  284 
    &  0  & 0.002 & -67  &  0.033  &  214  &  408 \\ 
    DASCMOP2  &  0.925 & 0.508 & -175  &  -0.035  &  267  &  385 
    &  0  & 0.002 & -31  &  -0.009  &  222  &  559 \\ 
    DASCMOP3  &  -0.004 & 0.015 & 1338  &  -0.038  &  328  &  73 
    &  0.001  & 0.007 & 423  &  0.005  &  263  &  87 \\ 
    DASCMOP4  &  0  & 0 & -1786  &  0.277  &  410  &  0 
    &  0  & 0 & 127  &  1.398  &  212  &  0 \\ 
    DASCMOP5  &  0.114 & 0.041 & -1793  &  0.358  &  403  &  0 
    &  0.114 & 0.041  & 120  &  1.148  &  214  &  0 \\ 
    DASCMOP6  &  0  & 0 & -1777  &  0.411  &  402  &  0 
    &  0  & 0 & 111  &  1.576  &  211  &  0 \\ 
    DASCMOP7  &  0  & 0 & -1908  &  1.812  &  420  &  0 
    &  0  & 0 & -69  &  0.993  &  229  &  0 \\ 
    DASCMOP8  &  0  & 0 & -1890  &  -0.364  &  448  &  0 
    &  0 & 0 & 2  &  -0.82  &  246  &  0 \\ 
    DASCMOP9  &  -0.023 & 0.407 &  3208  &  -0.208  &  363  &  13 
    &  -0.128 & 0.051 & 366  &  0.035  &  268  &  21 \\ 
   
    \rowcolor[gray]{.95}MOON  &  0.123  & 0.043 & -233  &  0.002  &  875  &  2 
     &  -0.103 & 0.432 & -16  &  -0.03  &  743  &  2 \\  
    \rowcolor[gray]{.95}MAZDA  &  0.007  & 0.026 & 1107  &  0.367  &  230  &  0 
     &  0.008  & 0.025 & -149  &  -0.472  &  209  &  0 \\  
   
    CRE21  &  0.002  & 0.009 & -3306  &  -0.005  &  797  &  0 
     &  0.007  & 0.014 & -314  &  0  &  860  &  0 \\  
    CRE22  &  0  & 0.001 & 158  &  0.028  &  445  &  14 
     &  0 & 0 & 85  &  0.064  &  380  &  17 \\  
    CRE23  &  0.005 & 0.009 & 969  &  0.013  &  1598  &  1 
     &  0.001  & 0.005 & 249  &  -0.013  &  1299  &  1 \\  
    CRE31  &  -0.001 & 0.003 & 224  &  0  &  444  &  7 
     &  0.002  & 0.002 & 164  &  -0.038  &  414  &  8 \\  
    CRE32  &  0.011 & 0.031 & 2767  &  0.033  &  745  &  0 
     &  -0.003 & 0.013 & 950  &  0.076  &  607  &  0 \\

    \rowcolor[gray]{.75} &\multicolumn{6}{c|}{\textbf{Operators variant}} & \multicolumn{6}{c}{\textbf{No-restart variant}}\\

    DASCMOP1  &  0.001  & 0.002 & 1389  &  -0.109  &  216  &  461 
    &  0.226  & 0.474 & 3739  &  0.038  &  427  &  276 \\ 
    DASCMOP2  &  -0.001 & 0.002 & 1129  &  0.004  &  223  &  789 
    &  0.004  & 0.003 & 3893  &  0.004  &  416  &  356 \\  
    DASCMOP3  &  0.005  & 0.005 & 41278  &  0.07  &  266  &  79 
    &  0.007  & 0.004 & 4804  &  0.043  &  782  &  54 \\  
    DASCMOP4  &  -0.357  & 0.574 & -361  &  1.549  &  215  &  0 
    &  -0.233  & 0.490 & 2142  &  0.352  &  168  &  0 \\  
    DASCMOP5  &  -0.245  & 0.578 & -291  &  1.488  &  217  &  0 
    &  0  & 0.363 & 2170  &  0.947  &  174  &  0 \\  
    DASCMOP6  &  -0.36  & 0.58 & -273  &  1.09  &  215  &  0 
    &  0  & 0 & 2134  &  1.032  &  172  &  0 \\  
    DASCMOP7  &  -0.656  & 0.691 & -471  &  1.454  &  220  &  0 
    &  -0.124 & 0.392 & 2241  &  0.345  &  135  &  0 \\  
    DASCMOP8  &  -0.787 & 0.668 & -537  &  0.311  &  234  &  0 
    &  -0.122 & 0.384 & 2304  &  -0.068  &  143  &  0 \\  
    DASCMOP9  &  -0.126 & 0.013 & 1618  &  0.048  &  266  &  16 
    &  -0.014 & 0.402 & 11593  &  -0.124  &  2202  &  18 \\ 
   
    \rowcolor[gray]{.95}MOON  &  -0.163  & 0.364 & 135  &  -0.026  &  727  &  2 
     &  0.307  & 0.379 & 1267  &  -0.011  &  108  &  2 \\  
    \rowcolor[gray]{.95}MAZDA  &  0.005  & 0.024 & -1323  &  1.288  &  237  &  0 
     &  0.01 & 0.029 & 1113  &  -0.709  &  906  &  4 \\  
    
    CRE21  &  0.017 & 0.040 & 643  &  0.054  &  787  &  0 
     &  0.27 & 0.423 & 4318  &  0.051  &  416  &  0 \\  
    CRE22  &  0 & 0 & 96  &  0.053  &  393  &  17 
     &  0 & 0.001 & 1849  &  0.079  &  157  &  14 \\  
    CRE23  &  0 & 0.002 & -2440  &  0.002  &  1478  &  1 
     &  -0.053 & 0.003 & 3186  &  0.005  &  1336  &  1 \\  
    CRE31  &  -0.002 & 0.001 & -22  &  0.001  &  447  &  8 
     &  -0.002 & 0.004 & 3546  &  -0.038  &  619  &  2 \\  
    CRE32  &  0 & 0.012 & 850  &  -0.002  &  665  &  0 
     &  -0.003 & 0.016 & 7569  &  0.006  &  2310  &  0 \\

    \rowcolor[gray]{.75} &\multicolumn{6}{c|}{\textbf{No-RA  variant}} & \multicolumn{6}{c}{\textbf{}}\\
    
    DASCMOP1  &  0 & 0.002 & -37  &  -0.061  &  13  &  449 &&&&& \\
    DASCMOP2  &  -0.001 & 0.002 & 240  &  0.005  &  12  &  537 &&&&& \\
    DASCMOP3  &  0.005 & 0.006 & -75  &  0.012  &  15  &  70 &&&&& \\
    DASCMOP4  &  0 & 0 & -108  &  0.586  &  11  &  0 &&&&& \\
    DASCMOP5  &  0.114 & 0.041 & -45  &  -0.014  &  13  &  0 &&&&& \\
    DASCMOP6  &  0 & 0 & -97  &  0.018  &  12  &  0 &&&&& \\
    DASCMOP7  &  -0.121 & 0.381 & -58  &  0.51  &  14  &  0 &&&&& \\
    DASCMOP8  &  -0.118 & 0.374 & -49  &  -0.52  &  15  &  0 &&&&& \\
    DASCMOP9  &  0.008 & 0.395 & 493  &  -0.093  &  12  &  14 &&&&& \\
   
    \rowcolor[gray]{.95}MOON  &  -0.052 & 0.406 & -14  &  -0.052  &  417  &  2 &&&&& \\
    \rowcolor[gray]{.95}MAZDA  &  0.006 & 0.03 & -141  &  0.435  &  12  &  0  &&&&& \\
   
    CRE21  &  0.004 & 0.01 & 78  &  0.065  &  550  &  0 &&&&& \\
    CRE22  &  0 & 0 & -55  &  0.053  &  35  &  24 &&&&& \\
    CRE23  &  -0.013 & 0.002 & -286  &  -0.013  &  622  &  2 &&&&& \\
    CRE31  &  -0.001 & 0.002 & 53  &  -0.004  &  24  &  8 &&&&& \\
    CRE32  &  -0.002 & 0.017 & 1 36  &  -0.001  &  39  &  0   &&&&& \\

\end{tabular}
\end{table*}

The different metrics for the variants are shown in Table~\ref{metrics_all}. Given that we already discussed the HV values above, we focus here on the other metrics shown in this Table. 

For the DASCMOP problems, all auto-MOEA/D variants find solutions in the Pareto front, suggesting that reaching the theoretical Pareto front of these problems is not a challenge. The update and the operators variants generally have a lower number of nodes and final population variance. Interestingly the aggregation function variant seems to have little effect in most of the metrics, and we highlight that the number of shared nodes is one or two orders of magnitude higher than the other variants for the DASCMOP3-9. Changing the decomposition method and the population size leads to more differences in areas of the decision space explored but doesn't affect much the overall HV performance. The same is true for the no-RA variant. This indicates that changing how MOEA/D works with the population during the search can affect how the decision space is explored but doesn't correspond to increments in HV performance for these problems. 

However, for the simulated and analytical real-world problems, we can observe little similarities among the five problems analyzed. This suggests that, unlike the DASCMOP set, the features of each of the simulated and analytical problems impact the auto-MOEA/D differently. For the MOON problem, we can see that the no-restart variant explored fewer areas of decision space, as we can see by the higher difference in the number of nodes. The number of shared nodes is the lowest among all variants, which suggests that auto-MOEA/D and the no-restart variant visit different areas of the decision space. That said, this variant was still able to find solutions in the approximated Pareto front. This suggests that the initial population has a high impact on the search exploration and all methods can follow a path to optimal solutions. We understand that this problem could be seen as a multimodal problem with at least one funnel to optimal solutions; however, more work needs to be done to validate this. 

Now, moving to the results of the MAZDA problem. For this problem, the variant that leads to less number of shared nodes, with only 5, the highest number of nodes and the highest final population variance is the decomposition+pop. size variant. This indicates that this variant can visit more areas of the decision space that are not explored by auto-MOEA/D while also having the final population spread to many different areas (given the higher variance). This could mean that choosing the right decomposition method and population size is critical for this problem. To our surprise, this is the only problem where no-restart leads to a noteworthy amount of solutions in the approximation to the Pareto front. Overall, these results show that these problems might have a set of unique characteristics in comparison to the other problems studied here.

Finally, the results of the CRE problems show that in terms of the number of nodes the no-restart variant is the one that has the biggest differences, exploring less the decision space and that not using RA was able to increase the number of solutions in the approximation to the Pareto front, with little differences in the other metrics. That is, the aggregation function has generally the highest number of shared solutions in the two sets of problems and the no-restart variant has about the same difference in terms of the number of nodes. This is in agreement with Table~\ref{metrics_ablation}. Thus we believe that DASCMOP1-3 and 9 might share similar problems characteristics with the CRE problem set.

\subsection{STNs Extension for Pairs of MOEAs}

For creating merged STN models of pairs of MOEAs, we first need to create one STN for each algorithm. To create the STN of a single algorithm, we follow a recently proposed methodology \cite{STN_MOP_evostar, lavinas_gecco2022}. 

As discussed (Section~\ref{prelim}), we extend this approach by merging the trajectories of two of these STNs by joining the two STNs graphs. This merged STN model contains the nodes and edges present in the STN of at least one algorithm. Attributes are kept for the nodes and edges, indicating whether they were visited by both algorithms (shared) or by one of them only. 


\begin{figure*}[htbp]
\centering
    \begin{subfigure}[!t]{0.49\textwidth}
    	\includegraphics[width=1\textwidth]{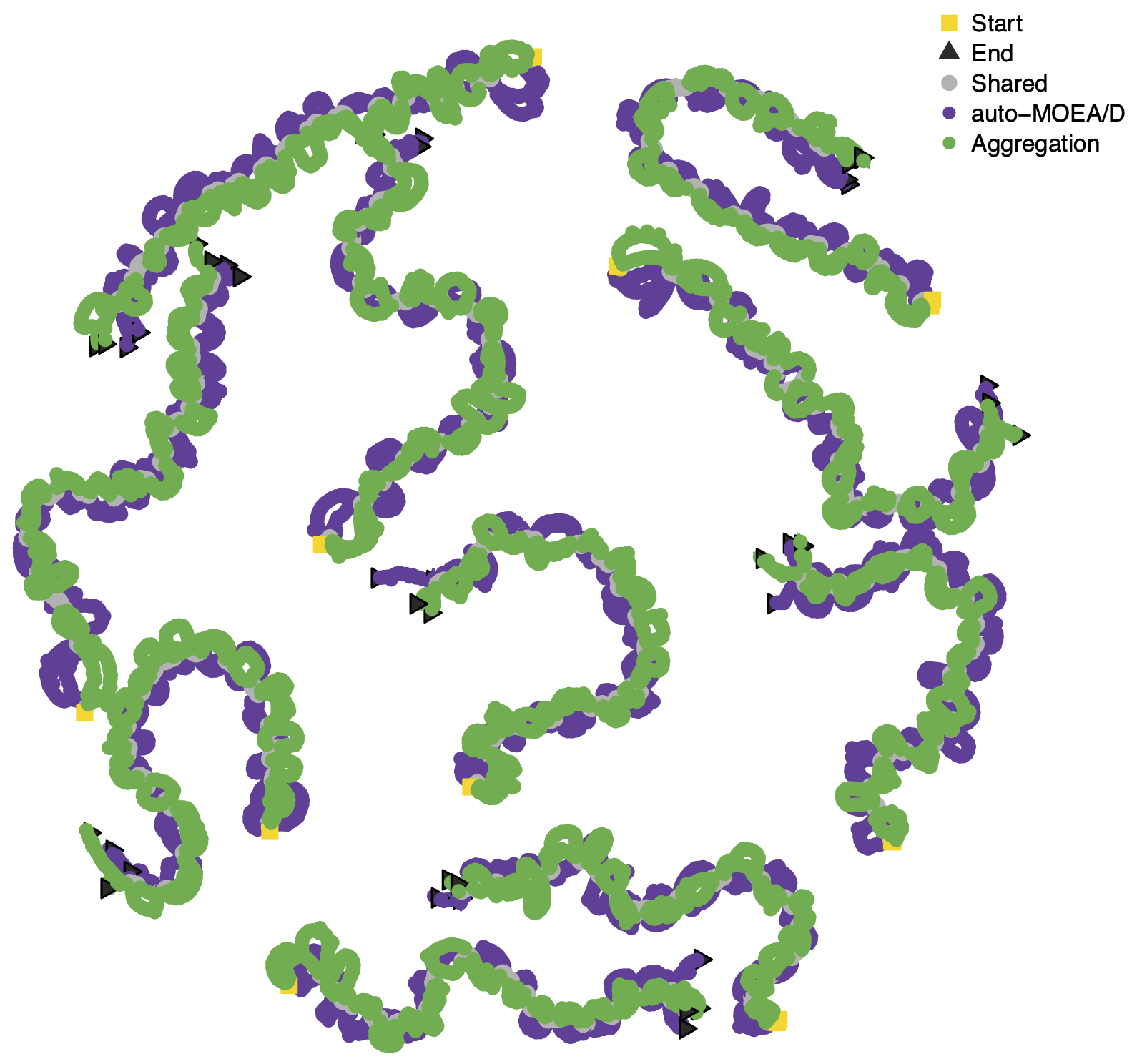}
    	\caption{DASCMOP1}
	\end{subfigure}
	~~
    \begin{subfigure}[!t]{0.49\textwidth}
    	\includegraphics[width=1\textwidth]{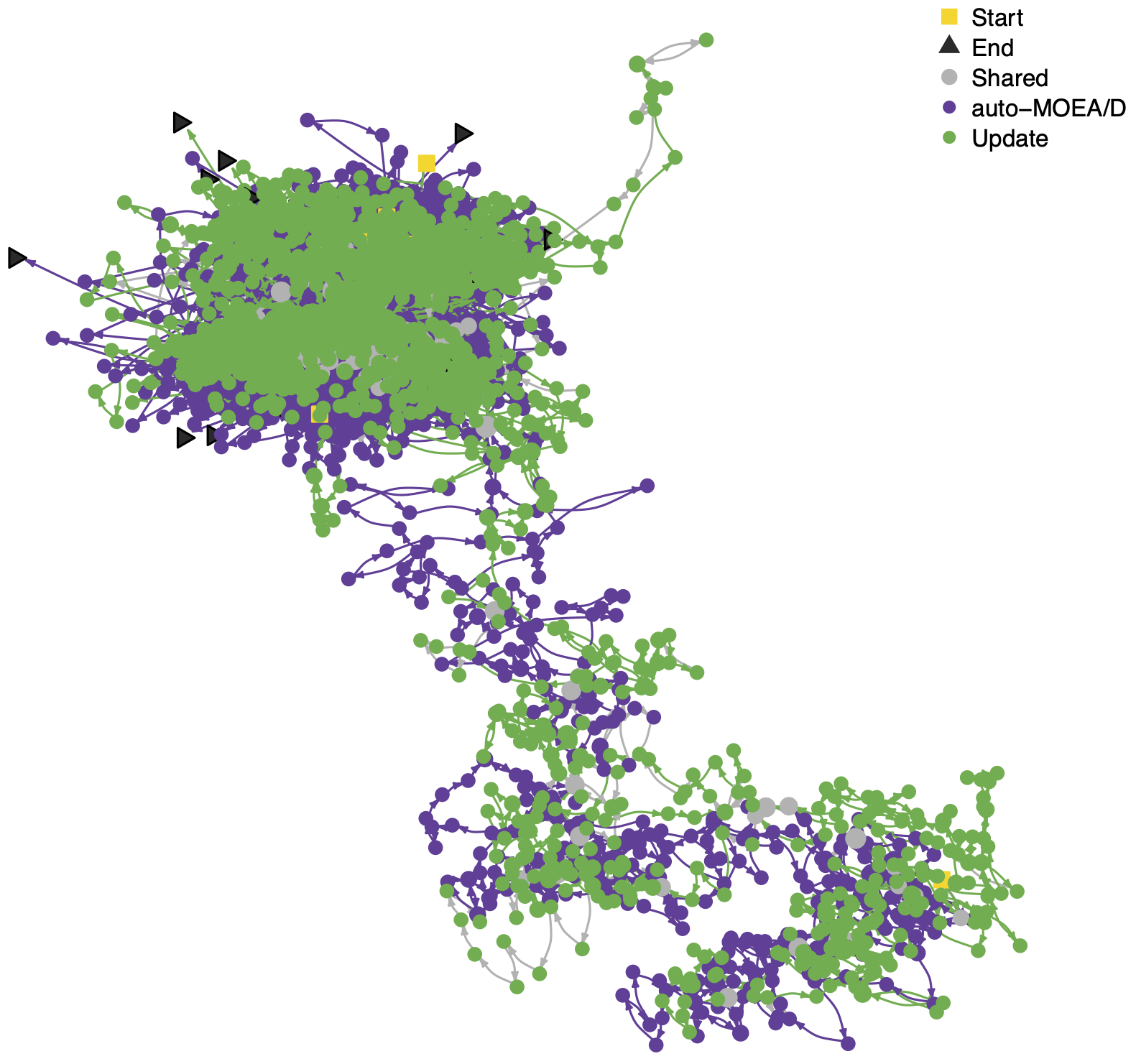}
    	\caption{DASCMOP2}
	\end{subfigure}
	
	\begin{subfigure}[!t]{0.49\textwidth}
    	\includegraphics[width=1\textwidth]{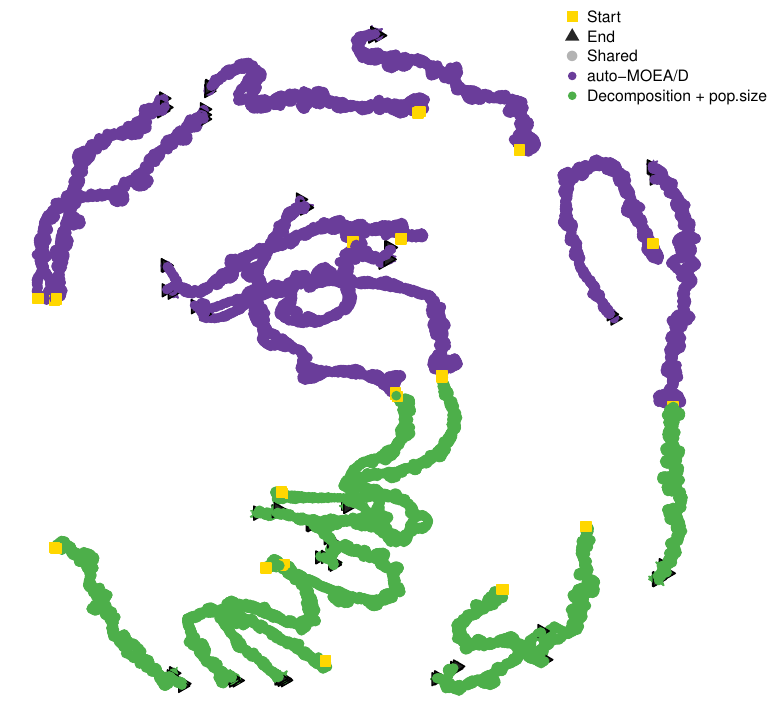}
    	\caption{DASCMOP3}
	\end{subfigure}
	~~
    \begin{subfigure}[!t]{0.49\textwidth}
    	\includegraphics[width=1\textwidth]{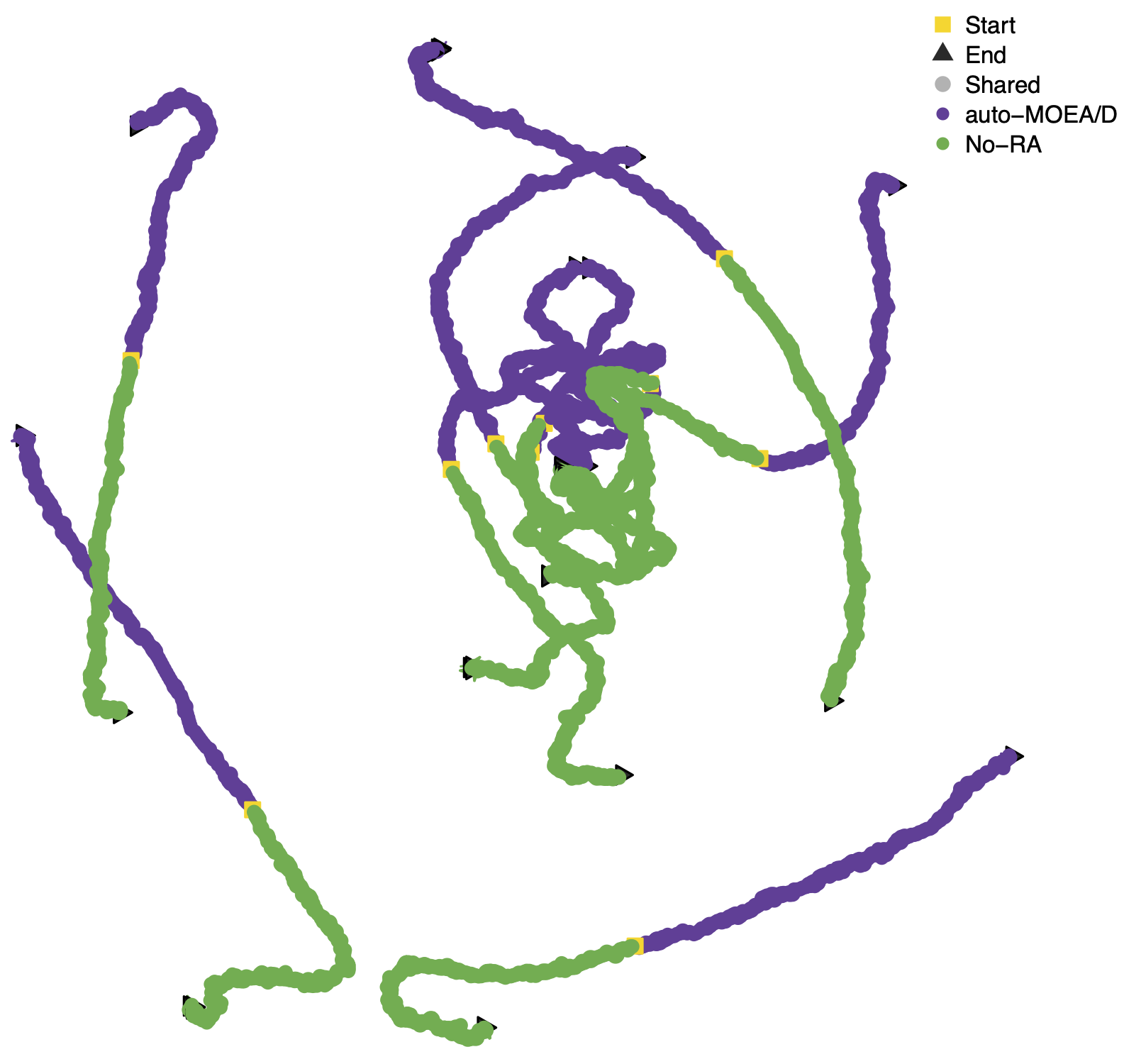}
    	\caption{DASCMOP4}
	\end{subfigure}
	 \caption{STN of auto-MOEA/D and different variants on the easy DASCMOP group. We see a diverse set of behaviors, from interlinked trajectories on the upper side to trajectories that visit distinct regions of the decision space on the bottom side.}
    \label{fig:stn_DASCMOP}
\end{figure*}

\begin{figure*}[htbp]
\centering
    \begin{subfigure}[!t]{0.49\textwidth}
    	\includegraphics[width=1\textwidth]{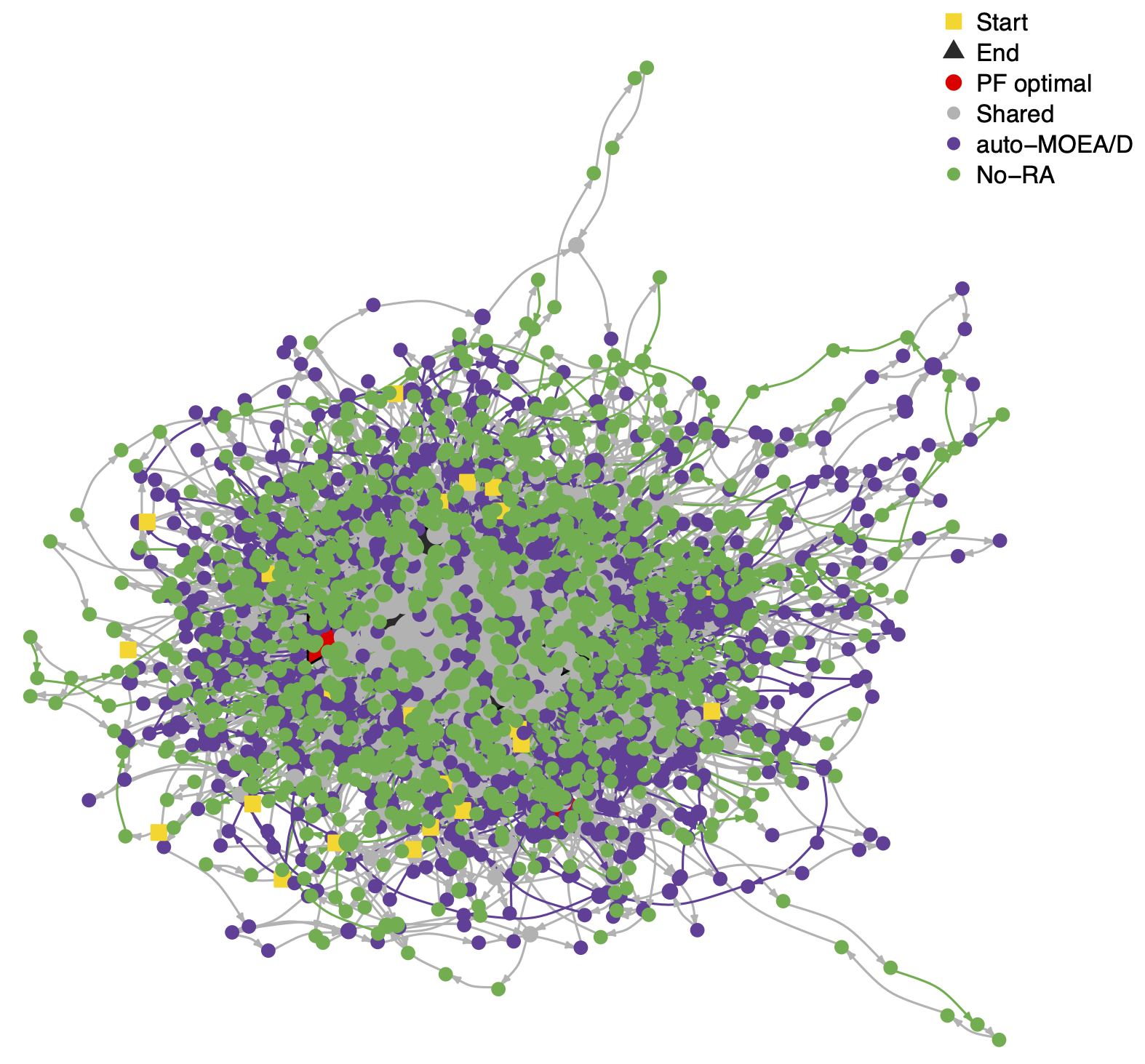}
	\caption{The trajectories of auto-MOEA/D and the no-RA variant overlap and share multiple locations, as shown by the grey nodes.}
	\end{subfigure}
	~~
    \begin{subfigure}[!t]{0.49\textwidth}
    	\includegraphics[width=1\textwidth]{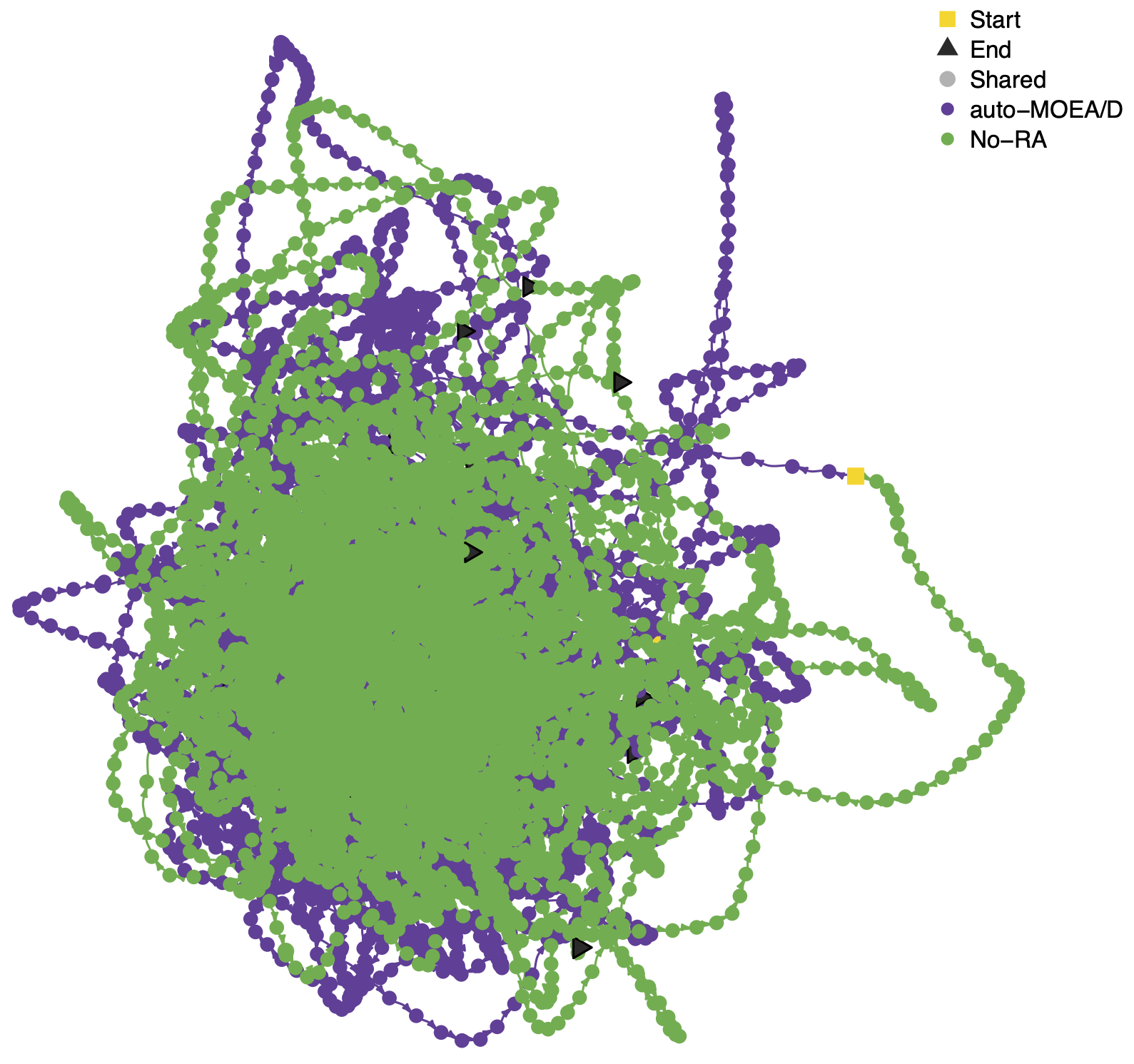}
	\caption{The trajectories of auto-MOEA/D and the no-RA variant share few locations (not visible).}
	\end{subfigure}
     \caption{STNs of auto-MOEA/D and two variants on the MOON (left) and MAZDA (right) problems. The MAZDA problem has a bigger effects the trajectories of the variants.}
    \label{fig:stn_simulated}
\end{figure*}

\begin{figure*}[htbp]
\centering
    \begin{subfigure}[!t]{0.49\textwidth}
    	\includegraphics[width=1\textwidth]{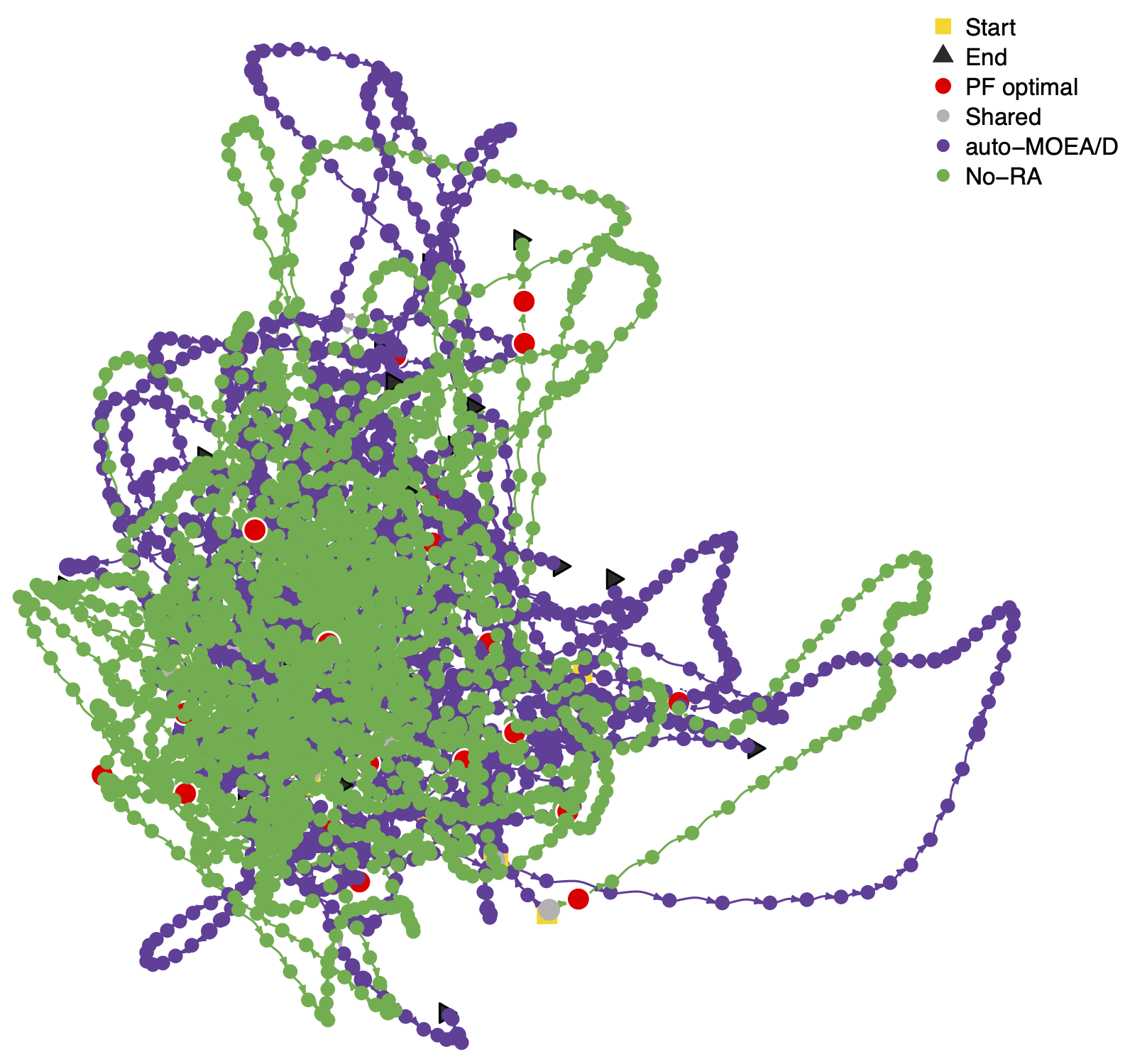}
	\caption{The trajectories of auto-MOEA/D and the no-RA variant visit many optimal solutions.}
	\end{subfigure}
	~~
	\begin{subfigure}[!t]{0.49\textwidth}
        \includegraphics[width=1\textwidth]{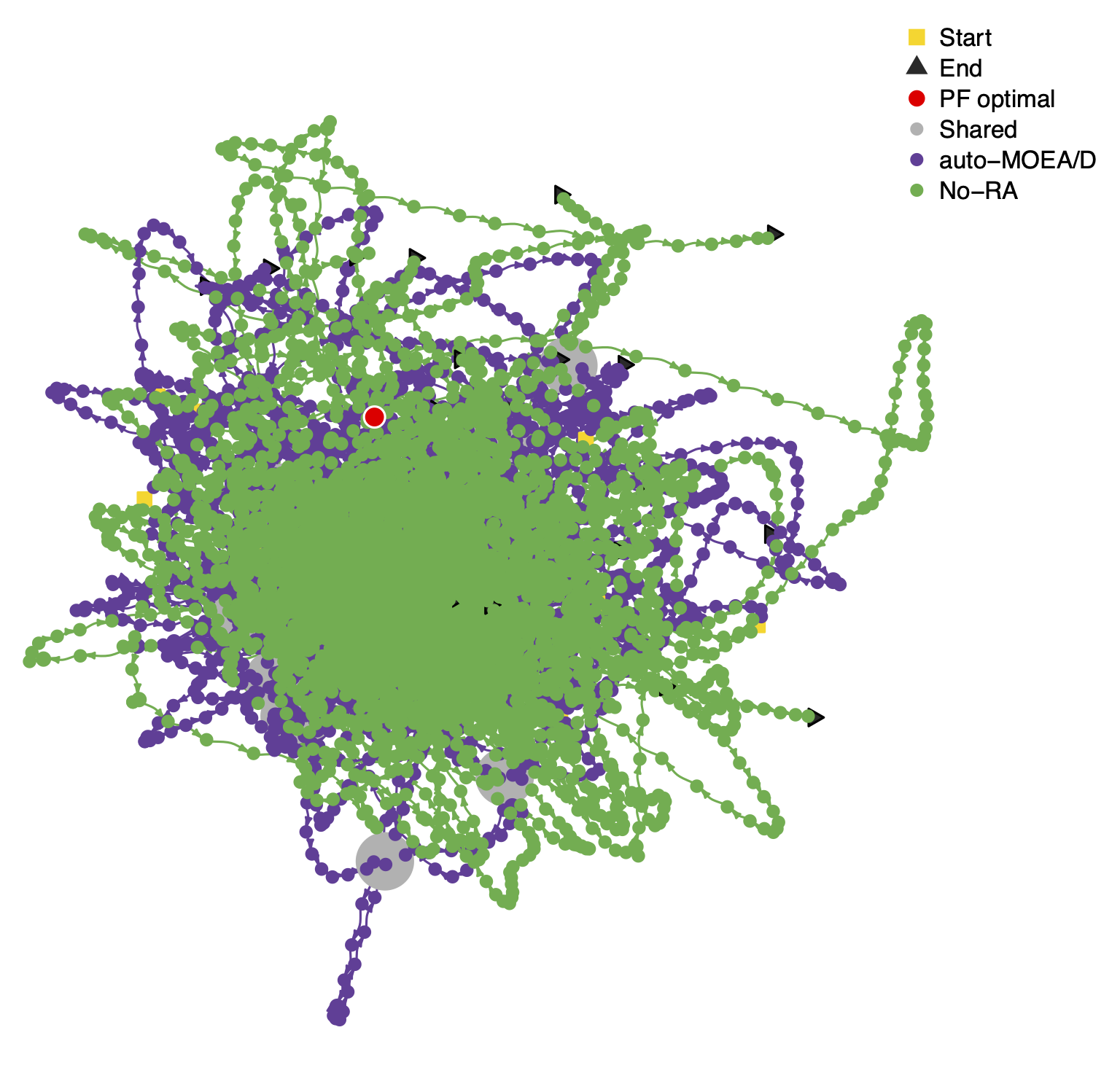}
    	\caption{The trajectories of auto-MOEA/D and the no-RA variant visit few optimal solutions.}
    	\end{subfigure}
     \caption{STNs of auto-MOEA/D and two variants on the CRE22 (left) and CRE23 (right) problems. The variants reach the approximation to the Pareto front.}
    \label{fig:stn_CRE_2obj}
\end{figure*}

\begin{figure*}[htbp]
\centering
    \begin{subfigure}[!t]{0.5\textwidth}
    	\includegraphics[width=1\textwidth]{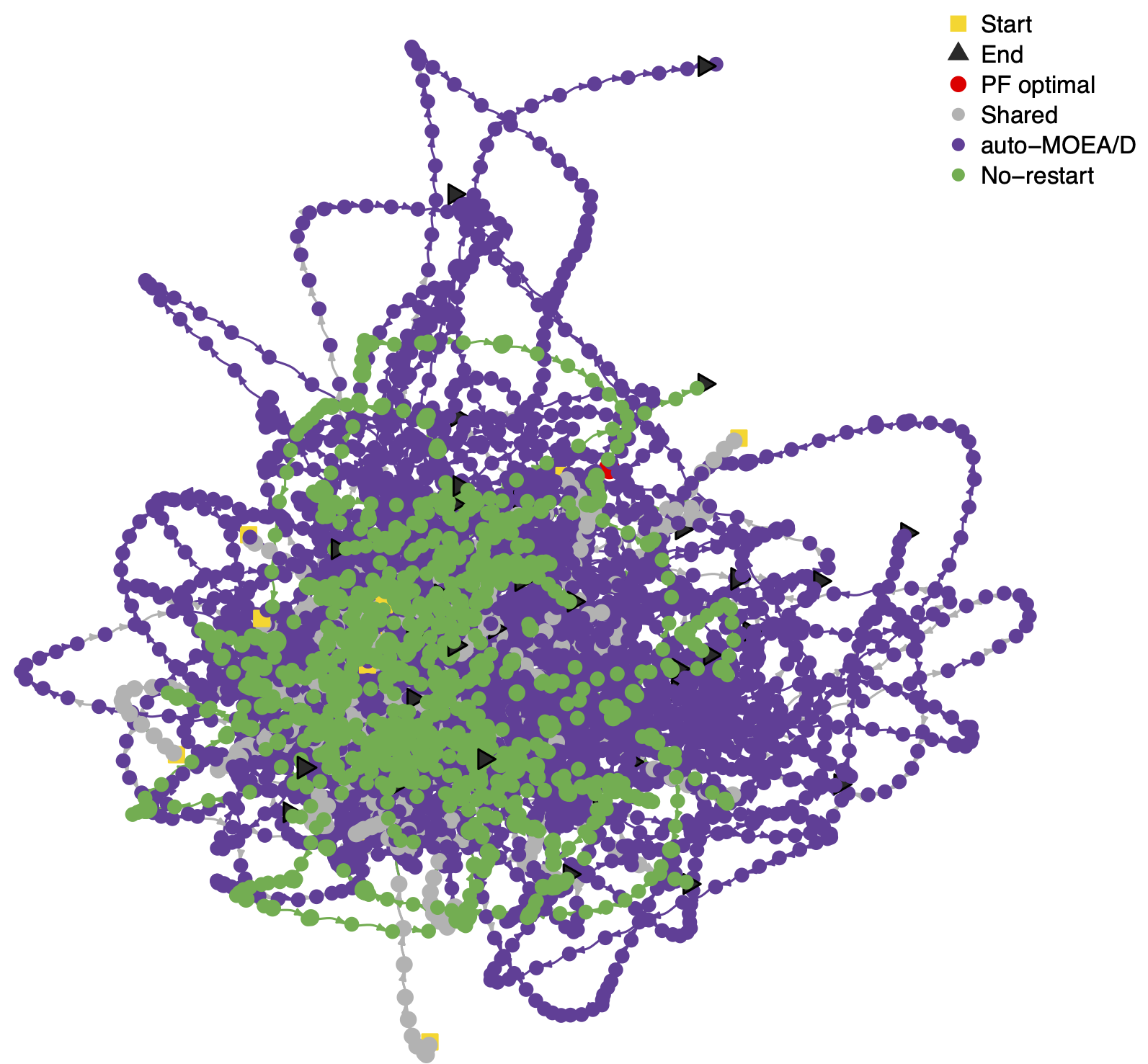}
	\caption{The trajectories of auto-MOEA/D and the operators variant are similar.}
	\end{subfigure}
	~~
    \begin{subfigure}[!t]{0.5\textwidth}
    	\includegraphics[width=1\textwidth]{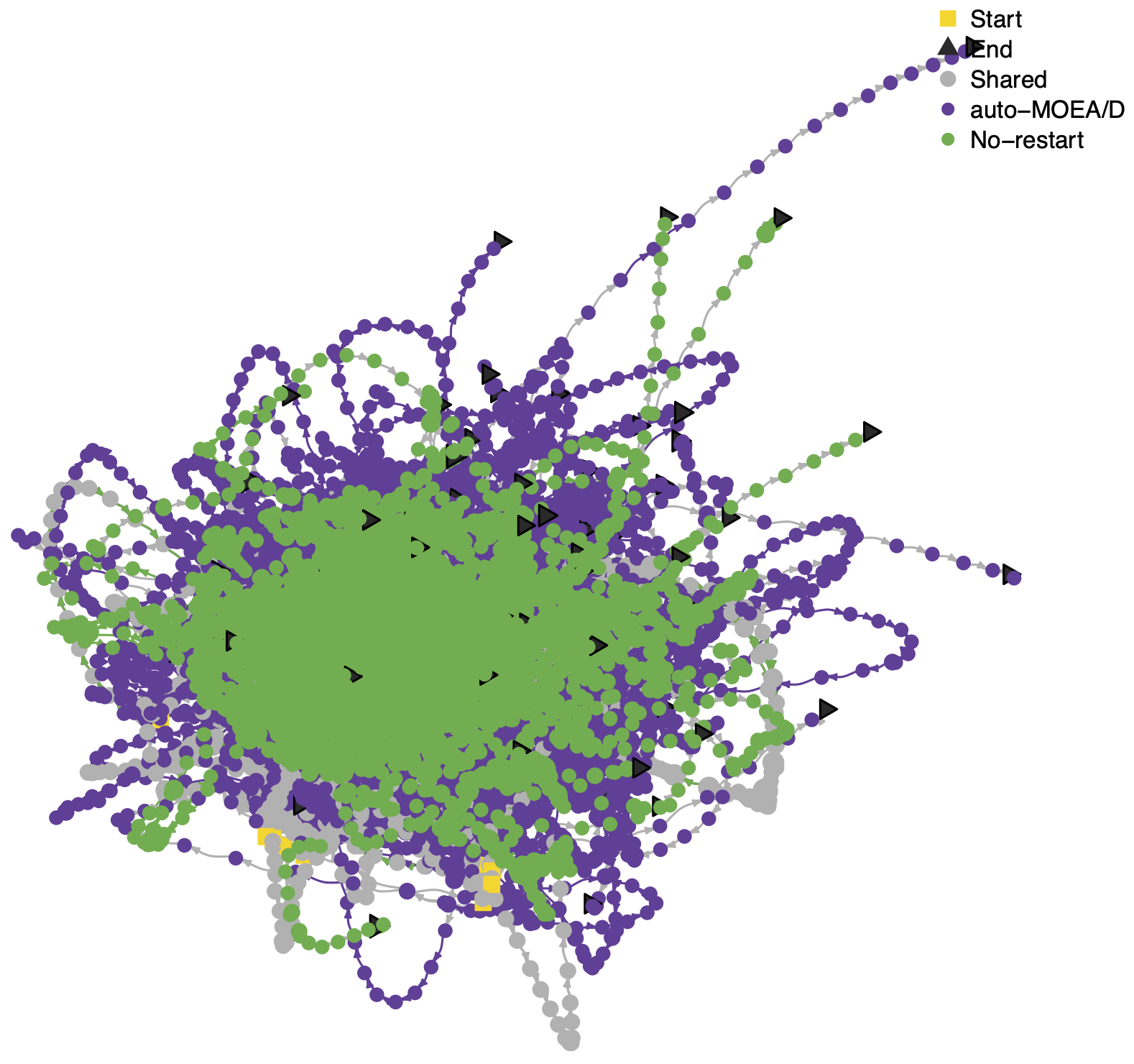}
	\caption{The trajectories of auto-MOEA/D and the no-restart variant share some nodes.}
	\end{subfigure}
	\vspace{1em}
     \caption{STNs of auto-MOEA/D and the two variants on the CRE31 (left) and CRE22 (right) problem.}
    \label{fig:stn_CRE_3obj}
\end{figure*}

Moving on to the STNs visualisations, Figures~\ref{fig:stn_DASCMOP},~\ref{fig:stn_simulated}, ~\ref{fig:stn_CRE_2obj} and ~\ref{fig:stn_CRE_3obj}. We selected visualisations of a representative variant for each problem. Considering the colours used in the STN visualisations, yellow
squares indicate the start of trajectories, and black triangles indicate the end of trajectories. The red colour shows the Pareto optimal solutions, and light grey circles show shared locations visited by both algorithms in that MOP. Finally, the trajectories of each algorithm are shown in different colours: purple for the auto-MOEA/D and green for the variant.

First of all, we comment on the overall differences among the STNs of the DASCMOP problems, Figure~\ref{fig:stn_DASCMOP}, and the simulated and analytical real-world problems ~\ref{fig:stn_simulated}, ~\ref{fig:stn_CRE_2obj} and ~\ref{fig:stn_CRE_3obj}. Unlikely the anytime HV performance that showed a similar objective space behavior between the DASCMOP and CRE problems, we can see that the decision space behavior of the DASCMOP STNs is more diverse when compared to the behavior of shown by the STNs for the real-world problems. This finding demonstrates how essential it is to explore the decision space behavior of algorithms. 

For the simulated MOON and MAZDA problem in Figures~\ref{fig:stn_simulated}, we can see that the trajectories are similar, with multiple shared nodes. That is, the trajectories of the STNs of the auto-MOEA/D and each variant overlap, visiting similar regions in the decision space. We associate this behavior with the number of shared nodes for this problem being high for all variants. Although we only showed the STNs of one pair, an identical trend occurs for all pairs of the auto-MOEA/D variants for the MOON problem. We understand this is indicative that the features of this simulated problem affect all MOEAs studied here in a similar fashion. The opposite happens for the MAZDA problem, where the trajectories shown in the Figure visit unrelated areas of the decision space.  For the pair selected, the trajectories do not overlap, and the number of shared nodes is much smaller. However, we can see in Table~\ref{metrics_all} that there is no agreement among the different variants in the metrics indicating a high impact of the problem difficulty on the search behavior of the different variants exploring the problem differently.

Moving now to the analytical CRE problems in Figures~\ref{fig:stn_CRE_2obj} and ~\ref{fig:stn_CRE_3obj}. We highlight that the trajectories of all variants for these problems overlap each other. Thus, we understand that the auto-MOEA/D variants visit similar regions in the decision space. Although the number of shared nodes is high for most of the variants in these problems, we can see that the variants are affected differently given the problem in question, indicating a contrasting set of features among the problems. In terms of best Pareto optimal solutions, we can see that there are much more solutions in the approximation to the Pareto front for the CRE22 problem than for the CRE23. We can see in Table~\ref{metrics_all} that an identical trend occurs for all pairs of the auto-MOEA/D variants for both problems. For the three objective problems, CRE31 and CRE32 we can see that the number of Pareto optimal solutions reduce substantially.

\section{Conclusion}
\label{discussion}

This works defines a new methodology to investigate the effects of algorithmic components based on our previous work~\cite{lavinas_gecco2022} that takes into account the decision space dynamics, as well as the objective space dynamics of the components. This methodology allows the user to investigate the impacts of the components of multiobjective algorithms in analytical and simulated problems with constraints. We contrasted the behavior of these configurations in terms of how they explore the decision space by comparing their Search Trajectory Networks (STNs), their population diversity, and how they behave in the objective space by exploring the anytime performance effects in terms of HV values. This analysis allowed us to identify the most influential components in the different problems we studied here.

Interestingly, the results shown in this paper differ from the ones found on our previous work. This is mainly because the training method used in the original paper was using a set of constraints that were too hard for MOEA/D to deal with. Since we wanted to improve the overall quality of the tuned MOEA/D in the hard set of problems we used a set of constraint difficulty triples of the DASCMOP benchmark set. That lead to increments in the performance of the tuned MOEA/D in the problems it was already performing well, but not affecting the performance in problems it had difficulties with. Thus, we understand that the results and conclusions are not accidental, but dependent on the different algorithm configuration selected.

We applied this methodology the auto-MOEA/D, a tuned MOEA/D designed by the irace package, and the subsequently derived variants that differ from this machine-designed MOEA by a single component. Our results showed that the most potentially influential variants differ given the set of problems: (1) for the DASCMOP problems the update variant showed more different behaviors in the objective and decision spaces; (2) the no-restart variant was more affected by the features of MOON problem while the decomposition+pop. size variant was the one that was more affected by the features of the MAZDA problem and (3) for the CRE family the variants that caused more changes in the decision space exploration behavior are the aggregation function and the no-restart.  However, it is still necessary to establish the generalization of such results ans how can we extrapolate to characterize the behavior of components independently of the scenarios observed.

We found that analysing the objective and the decision space simultaneously provides complementary information about how algorithms behave as the search progresses. In addition, the decision space behavior analysis was able to slightly contribute to the characterization of problems. For example, given how  the trajectories of the STNs of the auto-MOEA/D and each other variant for the MOON problem frequently overlap, thus this problem could be seen as a multimodal problem with at least one funnel to optimal solutions; but, more work needs to be done to validate this. Moreover, this finding demonstrates how essential it is to explore the decision space behavior of algorithms. 

In summary, this study strengthens the view that characterizing the effects of MOEA/D algorithm components could help in developing even more effective MOEAs. Taken together, these findings suggest a role for improving in promoting the study of specific components to develop new and better components. We understand that our results are of interest to the broad multiobjective evolutionary computation community. 

One limitation of this methodology is that the search space of possible algorithm configurations is limited by the choices of components and components parameters. However, with a careful selection of those, automatic composition can be a powerful tool to explore this possibility space.

\clearpage

\bibliographystyle{ACM-Reference-Format}
\bibliography{references}

\end{document}